\pdfoutput=1

\documentclass[11pt]{article}

\usepackage[]{acl}

\usepackage{times}
\usepackage{latexsym}
\usepackage{colortbl}
\usepackage{bbding}
\usepackage{pdfpages}

\usepackage[T1]{fontenc}

\usepackage[utf8]{inputenc}

\usepackage{microtype}

\usepackage{inconsolata}

\usepackage{graphicx}

\usepackage{booktabs}
\usepackage[intlimits]{amsmath}  
\usepackage[colorinlistoftodos,size=tiny,disable]{todonotes}

\usepackage{caption}
\usepackage{pifont}
\usepackage{subcaption}
\usepackage{multirow}
\usepackage{enumitem}
\usepackage{makecell}
\usepackage{tabularx}
\usepackage{soul}
\usepackage{amssymb}
\usepackage[rgb,dvipsnames]{xcolor}
\usepackage{afterpage}
\usepackage{longtable}
\usepackage{array}
\usepackage{ragged2e}
\usepackage{supertabular}
\usepackage[export]{adjustbox}
\usepackage[most]{tcolorbox}

\newcommand{\se}[1]{\textcolor{blue}{#1}}
\newcommand{\yc}[1]{\textcolor{black}{#1}}
\newcommand{\ycr}[1]{\textcolor{black}{#1}}

\newtcolorbox{definition}{
  enhanced,
  colback=blue!5!white, 
  colframe=blue!75!black, 
  coltitle=blue!90!black, 
  fonttitle=\bfseries,
  title=Definition,
  sharp corners,
  borderline west={3pt}{0pt}{blue!75!black},
  attach title to upper,
  after title={:\enskip},
  colupper=black 
}
\usepackage{listings}
\usepackage{arydshln}

\lstdefinestyle{promptstyle}{
    basicstyle=\scriptsize\ttfamily,
    breaklines=true,
    breakatwhitespace=true,
    frame=single,
    backgroundcolor=\color{gray!5}, 
    showstringspaces=false,
    columns=fullflexible,
    keepspaces=true
}

\newcommand{\fit}[0]{\textsc{Emo}$_{\text{fit}}$}
\newcommand{\evi}[0]{\textsc{Evi}$_{\text{prop}}$}
\newcommand{\norm}[0]{\textsc{Norm}$_{\text{acc}}$}
\newcommand{\logic}[0]{\textsc{Logic}$_{\text{sup}}$}
\newcommand{\eat}[0]{\textsc{EA}$_{\text{type}}$}
\newcommand{\eas}[0]{\textsc{EA}$_{\text{strength}}$}

\title{Justification of Emotional Appeal in Computational Argumentation}

\author{Yanran Chen and Steffen Eger \\
 Natural Language Learning and Generation Lab (NLLG) \\ \url{https://nl2g.github.io/} \\
University of Technology Nuremberg, Germany 
}

\begin{document}

\maketitle

\begin{abstract}
Emotional appeals (EAs) are often treated as fallacious in computational argumentation, yet they may be justified when warranted by the context. We introduce the task of predicting EA justification and present the first human-annotated dataset for this task, covering 1,176 argument--context pairs. Grounded in philosophical literature, our framework evaluates EA justification along logical and emotional dimensions. Human annotations show that justification criteria form a shared latent judgment space, with evidential proportionality and normative acceptability most strongly associated with overall justification. We further find that EAs without logical support are not always judged fully unjustified, and stronger EAs are not necessarily judged less justified. Finally, LLM experiments show that EA justification prediction remains challenging.
\end{abstract}

\section{Introduction}

\yc{
Emotional appeals (EAs) refer to attempts to evoke emotions in an audience
to support a claim, and belong to the \emph{pathos} dimension of
argumentation, alongside \emph{ethos} (speaker's credibility) and \emph{logos} (logic)
\citep{kennedy2006rhetoric}.
EAs have often been discussed as fallacies, but argumentation theory does
not consider them fallacious by default
\citep{brinton1988appeal,walton2010place,manolescu2006normative}.
Prior work argues that EAs can be reasonable when they are supported by the
context and used appropriately
\citep{brinton1988appeal,manolescu2006normative,
macagno2010argumentative}.
}

\yc{
In natural language processing (NLP), however, whether an EA is justified has received little attention.
Prior work mainly studies the relation between emotions and argument
convincingness
\citep{habernal-gurevych-2016-argument,
wachsmuth-etal-2017-computational,chen-eger-2025-emotions}.
EAs are also frequently modeled as fallacies or propaganda techniques
\citep{da-san-martino-etal-2019-fine,helwe-etal-2024-mafalda}.
Such formulations do not distinguish between EAs that are warranted by the
context and those that are not. For example, an appeal to fear may be
reasonable when there is sufficient evidence of serious danger, but
fallacious or manipulative when such evidence is absent.
}%
\yc{
This issue becomes more important as large language models (LLMs) make affective framing easy to
generate at scale: it can affect human fallacy detection \citep{chen-etal-2026-emotionally}, impair automated fake-news detection \citep{tahmasebi2026robustfakenewsdetection}, and shape readers' perception of news trustworthiness \citep{10.1145/3772318.3791527}. 
} 

\begin{figure}[t]
    \centering
    \includegraphics[width=\linewidth]{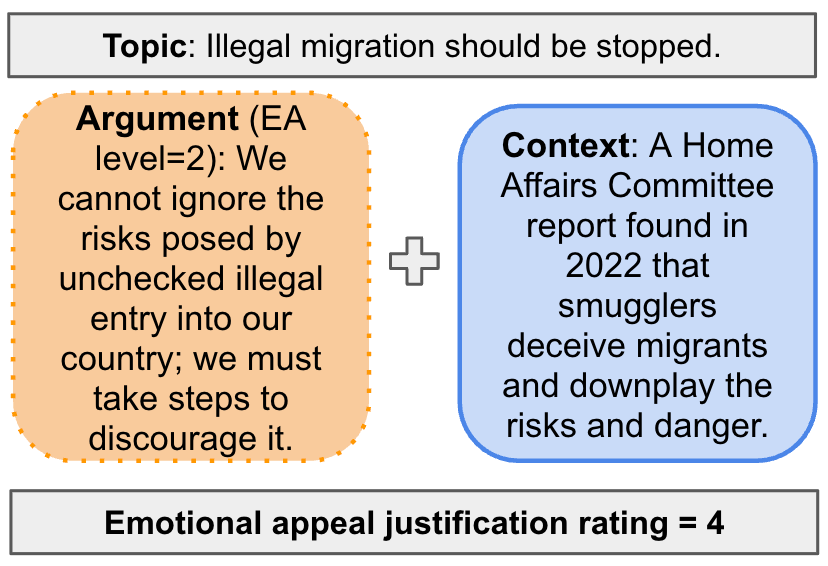}
    \caption{Example evaluation instance, where EA justification is rated for the argument with respect to the paired context and given topic; EA level = 2 and EA justification = 4.}
    \label{fig:eval_example}
    \vspace{-.4cm}
\end{figure}

To fill this gap, we propose the task of predicting EA justification, as illustrated in Figure~\ref{fig:eval_example}: given an emotional argument, the task is to evaluate its EA justification with respect to its context and topic. To the best of our knowledge, this is the first work to study EA justification in computational argumentation. We first define criteria for evaluating EA justification based on argumentation theory and philosophical work on emotional justification (\S\ref{sec:def}) and build an evaluation framework around them (\S\ref{sec:task}). We then collect human annotations on a curated argument dataset (\S\ref{sec:anno}). Following recent work that uses LLMs to generate synthetic or adversarial counterparts for controlled studies and data enrichment \citep{chen-eger-2025-emotions,chen-etal-2026-emotionally,tahmasebi2026robustfakenewsdetection}, we also augment real-world data with synthetic controlled variants (\S\ref{sec:syn}). Using these annotations, we analyze how humans perceive EA justification and how their judgments align with the theoretical criteria (\S\ref{sec:human}). Finally, we conduct a small-scale experiment to examine whether current LLMs can predict EA justification (\S\ref{sec:llm_predict}). Our contributions are three-fold:\footnote{Code + Data: \url{https://github.com/NL2G/EMCONA-UTN}}

\begin{itemize}[topsep=2pt,itemsep=-1pt,leftmargin=*]
    \item We propose the task of predicting EA justification for an argument, given its context and topic. We build an evaluation framework based on philosophical literature and collect human annotations for 1,176 argument--context pairs. 
    \item Using the collected data, we analyze how humans perceive EA justification. We find that the justification criteria share a latent judgment space rather than forming independent evaluation axes. Among these criteria, whether the EA strength is warranted by the context and whether the EA is socially and culturally acceptable are most strongly associated with overall justification. We also find that EAs without logical support are not always perceived as fully unjustified, and that stronger EAs are not necessarily perceived as less justified, partly complicating existing theories.
    \item We evaluate two LLMs, GPT-5.4-mini and DeepSeek-V3.2,
    on the EA justification prediction task. Their predictions correlate only weakly with human judgments, indicating both the difficulty of the task and current LLMs' limited ability to perform it, especially for emotion-related criteria. 
\end{itemize}

\section{Related Work}

This work connects to (1) 
computational argumentation and (2) emotional justification.

\paragraph{Computational argumentation}

\ycr{
Computational argumentation can be broadly divided into four types of tasks \citep{lauscher-etal-2022-scientia,ajjour-etal-2026-argbench}:
mining \citep[e.g.][]{eger-etal-2017-neural,cabessa-etal-2025-argument},
assessment \citep[e.g.][]{electronics13204088,jin-etal-2025-multi},
reasoning \citep[e.g.][]{habernal-etal-2018-name,jin-etal-2022-logical},
and generation \citep[e.g.][]{hua-etal-2019-argument-generation,altemeyer-etal-2025-argument}.
}

\yc{
EAs have mainly been studied in relation to argument convincingness and quality
\ycr{in the context of argument assessment}
\citep{wachsmuth-etal-2017-computational,ziegenbein-etal-2023-modeling,
chen-eger-2025-emotions,greschner2025trustmeiconvince,verma-etal-2026-predicting}.
They are also often treated as fallacies
\ycr{in argument reasoning}
\citep{da-san-martino-etal-2019-fine,
sahai-etal-2021-breaking,
jin-etal-2022-logical,
musi2022developing,
helwe-etal-2024-mafalda}.
More recently, LLMs have been used to generate or manipulate emotional
framing in arguments
\citep{chen-eger-2025-emotions,chen-etal-2026-emotionally},
showing both its effects on persuasion and its potential risks.
However, to our knowledge, no prior computational work has studied whether
an EA is justified in a given context.
}

\paragraph{Emotional justification}
\yc{
Philosophical work has examined when \textbf{emotions} are justified,
fitting, or rational. Emotions can be \emph{epistemically justified} when
they are reasonable given a person's evidence or knowledge
\citep{deonna2012justified,echeverri2019emotional,logins2024evidence},
and \emph{fitting} or \emph{appropriate} when they correctly represent
relevant evaluative features of a situation
\citep{Na’aman03072021}.
Their \emph{rationality} may further depend on a person's goals, their
interpretation of the situation, and relevant social expectations
\citep{scherer2011rationality}.
}

\yc{
Argumentation theory considers a related question: when is it justified for
an arguer to \textbf{appeal to emotion}?
\citet{brinton1988appeal} evaluates appeals to angry emotions by whether the emotion is well-grounded and proportionate. \citet{manolescu2006normative} argues that emotional appeals may be relevant and non-manipulative. \citet{macagno2010argumentative} hold that emotive language can reasonably support value-based argumentation when open to critical questioning. More recently, \citet{yu2026unpacking} defines an emotional appeal as fallacious when its conclusion appears justified but is not.
}

\yc{
\textbf{Overall}, prior work studies either the argumentative effects of EAs or the theoretical conditions under which they are justified. To the best of our knowledge, we are the first to connect these two lines of work by developing a feasible computational framework for studying EA justification.
}

\newcolumntype{Y}{>{\raggedright\arraybackslash}X}

\newcolumntype{P}[1]{>{\raggedright\arraybackslash}m{#1}}

\renewcommand{\tabularxcolumn}[1]{m{#1}}


\definecolor{eaHeader}{HTML}{DCEAF7}
\definecolor{eaBody}{HTML}{F3F8FC}

\definecolor{logicHeader}{HTML}{E4F1E4}
\definecolor{logicBody}{HTML}{F4FAF4}

\definecolor{emotionHeader}{HTML}{F4E5D5}
\definecolor{emotionBody}{HTML}{FCF7F1}

\definecolor{overallHeader}{HTML}{E9E1F3}
\definecolor{overallBody}{HTML}{F8F5FB}


\begin{table*}[t]
\centering
\small
\setlength{\tabcolsep}{4pt}
\renewcommand{\arraystretch}{1.12}

\begin{tabularx}{\textwidth}{@{}
P{0.20\textwidth}
P{0.42\textwidth}
Y
@{}}

\toprule
\textbf{Evaluation Criteria}
&
\textbf{Definition}
&
\textbf{Range}
\\

\midrule


\rowcolor{eaHeader}
\multicolumn{3}{@{}l}{
    \textit{Emotional Appeal}
}
\\

\midrule

\rowcolor{eaBody}
Type (\textbf{\eat})
&
The main emotion the argument seems intended to evoke as persuasive support.
&
unclear/no emotional appeal, fear/threat, pity/compassion/sympathy,
anger/indignation, hope, pride/flattery, guilt/shame, and loyalty/solidarity
\\
\hdashline

\rowcolor{eaBody}
Strength (\textbf{\eas})
&
How intense the emotional appeal is in the argument.
&
1 (no emotional appeal) -- 5 (very strong emotional appeal)
\\

\addlinespace[2pt]
\midrule


\rowcolor{logicHeader}
\multicolumn{3}{@{}l}{
    \textit{EA Justification -- Logical Dimension}
}
\\

\midrule

\rowcolor{logicBody}
Logical Support (\textbf{\logic})
&
To what degree the context logically supports or opposes the argument.
&
$-3$ (strongly opposes) -- 0 (neutral) -- $+3$ (strongly supporting)
\\

\midrule
\addlinespace[2pt]


\rowcolor{emotionHeader}
\multicolumn{3}{@{}l}{
    \textit{EA Justification -- Emotional Dimension}
}
\\

\midrule

\rowcolor{emotionBody}
Emotion-type Fit (\textbf{\fit})
&
To what degree the type of emotional appeal fits the situation described in the context.
&
1 (clearly no fit) -- 5 (very strong fit)
\\
\hdashline

\rowcolor{emotionBody}
Evidential Proportionality (\textbf{\evi})
&
To what degree the strength of the emotional appeal fits the severity of the situation evidenced by the context.
&
1 (clearly disproportionate) -- 5 (very strong proportionality)
\\
\hdashline

\rowcolor{emotionBody}
Reasons for reductions in \evi
&
The reasons for reductions in Evidential Proportionality.
&
No reduction / fully justified, opposing context, RIS, MAG, PROB,
TPROX, SPROX, and free text
\\
\hdashline

\rowcolor{emotionBody}
Normative Acceptability (\textbf{\norm})
&
To what degree the emotional appeal is acceptable under social and cultural norms given the context.
&
1 (clearly unacceptable) -- 5 (highly acceptable)
\\

\midrule
\addlinespace[2pt]


\rowcolor{overallHeader}
\multicolumn{3}{@{}l}{
    \textit{EA Justification -- Overall}
}
\\

\midrule

\rowcolor{overallBody}
Justification
&
How justified the emotional appeal in the argument is given the context.
&
1 (not justified at all) -- 5 (very strongly justified)
\\

\bottomrule

\end{tabularx}

\caption{
Evaluation framework for assessing the justification of emotional appeal
in an argument with respect to its context and topic.
}
\label{tab:ea-criteria}

\vspace{-.6cm}

\end{table*}
\section{Justification of Emotional Appeal}\label{sec:thory}

In this section, we first introduce our definitions of emotional justification in arguments, adapted from philosophical literature (\S\ref{sec:def}), and then present the evaluation framework designed for EA justification (\S\ref{sec:task}).

\subsection{Definitions}\label{sec:def}

\yc{A simple \textbf{argument} consists of one or more premises that provide support for a conclusion (or claim). \textbf{Emotional appeal} refers to an emotion intended to be evoked as persuasive support for a claim in an argument.}
If an EA is considered fallacious in an argument, it should naturally be unjustified; we therefore begin with the definition of fallacy. According to \citet{Hamblin1970-HAMF,hansen2023committing,yu2026unpacking}, fallacies have traditionally been defined as arguments that \emph{seem} valid but are not. Building on this view, \citet{yu2026unpacking} further defines an appeal to emotion as fallacious if ``\emph{its conclusion only appears to be justified but is not}.'' It can thus be evaluated by whether the conclusion is justified by the premises. Thus, we derive the first definition of emotional justification along the logic dimension of argumentation: \textbf{An emotional appeal is justified \emph{only if} the premises of the argument justify its conclusion} (\emph{Def.\ 1}).

\yc{
Prior work in argumentation theory suggests that EAs can be
justified when the emotion appealed to is appropriately grounded in the
argumentative context \citep{brinton1988appeal}; related philosophical work similarly evaluates emotions in
terms of their justification, appropriateness, and rationality \citep{scherer2011rationality,echeverri2019emotional,Na’aman03072021}.}
We therefore derive the second definition of EA justification:
\textbf{An emotional appeal is justified only \emph{if} \ycr{the} premise\ycr{s} \ycr{justify} the
emotion intended to appeal to} (\ycr{\emph{Def.~2}}).
We further define this justification using \textbf{three sub-criteria}:

\begin{itemize}[topsep=2pt,itemsep=-1pt,leftmargin=*]
    \item (1) \textbf{Emotion-Type Fit} (\fit{}): The type of emotional appeal (e.g., appeal to fear, appeal to pity) fits the situation described in the premise.
    \item (2) \textbf{Evidential Proportionality} (\evi{}): The strength of the emotional appeal fits the severity of the situation evidenced by the premise. For example, a strong appeal to fear may be less justified if the premise only provides evidence for a mildly dangerous situation. \citet{walton2010place,govier2010practical,yu2026unpacking} also emphasize the problem of strong emotional appeals. When the emotion type is justified by the premise, several subtle factors may affect this criterion. For example, an emotional appeal may be more justified when the premise cites evidence from an authoritative organization rather than an unknown source. By contrast, it may be less justified when its strength is exaggerated relative to the risk described in the premise.
    \item (3) \textbf{Normative Acceptability} (\norm{}): The emotional appeal is acceptable under social and cultural norms in the given context. We adapt this criterion from \citet{scherer2011rationality}, who argues that emotional rationality also depends on social norms, cultural values, and common expectations about appropriate responses in specific situations. 
\end{itemize}

Based on the above definitions, we design the following evaluation framework to study human-perceived EA justification and how it aligns with the literature.

\subsection{Evaluation Framework}\label{sec:task}

\paragraph{Task Formulation}
Evaluating EA justification first requires reconstructing the argumentative structure of the argument from a text, including the premises, conclusion/claim, and content that carries the EA. These argument mining steps are already challenging for untrained annotators. 
\yc{Moreover, EA can appear in any part of an argument: the claim/conclusion, the premise, or the non-argumentative part.}

To make the task more feasible, we formulate it as follows: \emph{Given an \textbf{argument--context pair} and its topic, we evaluate the EA justification of the argument with respect to the context}. An example is shown in Figure~\ref{fig:eval_example}. 
In this setup, \yc{we combine the EA-carrying content with the claim, so they together form an argument in which the EA functions as the only ``premise'' supporting the claim.} The context provides the information needed to interpret the situation and assess the EA justification. \yc{We use the term context instead of premise because, in real-world scenarios, relevant information may come from both the argument and external background knowledge. E.g., an appeal to pity for orphaned children may be justified by their circumstances, even if such information is not stated in the argument.}

\paragraph{Evaluation Criteria}
The evaluation criteria are summarized in Table~\ref{tab:ea-criteria}. To account for Def.~1, we first 
evaluate the degree to which the context \ycr{logically} supports the argument (\logic). After identifying the type and strength of the EA (\eat, \eas), we evaluate Def.~2 according to the three sub-criteria (\fit, \evi, \norm). Finally, we assign an overall justification score to each argument--context pair, indicating how justified the emotional appeal in the argument is given the context overall. We use absolute ratings rather than binary labels to capture fine-grained relations among these criteria. 

For \evi, we additionally annotate the reasons for score reductions to examine how humans relate \eas{} to these potentially influencing factors. We pre-define the following reasons: reliability of information source (RIS), magnitude of risks/outcomes (MAG), probability of risks/outcomes (PROB), spatial proximity (SPROX), and temporal proximity (TPROX). In addition, annotators can choose `No reduction / Fully justified' or `Opposing context'. Annotators may also write their own reasons if none of the pre-defined labels apply. We select these reasons based on examples in the literature and heuristics. For instance, \citet{echeverri2019emotional} provides \yc{examples} related to RIS: ``\emph{If Pat's fear that terrorists will attack New York City is based on CIA intelligence reports, her fear is justified.}'' and ``\emph{ If Peter is jealous of his wife on the basis of a mere hunch, his jealousy is unjustified.}''

\begin{figure*}[!t]
    \centering
    \includegraphics[width=.9\linewidth]{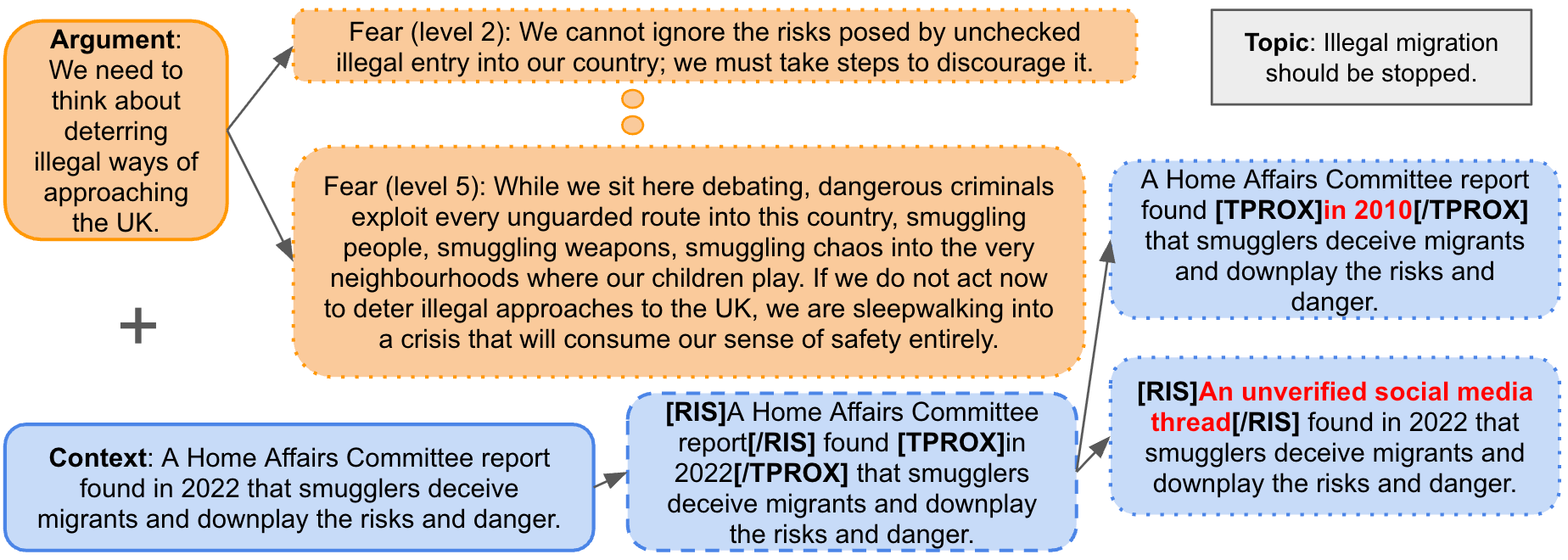}
    \vspace{-.3cm}
    \caption{Examples of emotional appeal injection and context perturbation. \ycr{Orange boxes represent arguments, and blue boxes represent contexts. Original argument--context pairs are shown with solid borders, while synthetic ones are shown with dashed borders.}}
    \label{fig:example}
    \vspace{-.3cm}
\end{figure*}

\section{Dataset Creation}\label{sec:data}
In this section, we describe how we create the annotation instances. Following prior work on controlled studies of affective effects, 
we use LLMs to generate synthetic counterparts (\S\ref{sec:syn}) for real-world data (\S\ref{sec:data_real}). We then conduct a human annotation on both the real-world and synthetic data according to our evaluation framework for EA justification (\S\ref{sec:anno}).

\subsection{Real-world Arguments \&  Contexts}\label{sec:data_real}

We first obtain noisy candidate argument--context pairs and their topics from \textbf{two data sources}: UK parliamentary debates (Hansard) and QAM \citep{guo-etal-2023-aqe}.\footnote{\yc{We call the raw pairs ``noisy'' because they were not directly usable: Hansard pairs were heuristically extracted from long debates by prompting LLMs and thus contain errors, while QAM pairs often suffer from imperfect sentence segmentation and sentence-level annotation granularity.}} 
QAM contains claim--evidence pairs for everyday debate topics, such as `Should we ban unsustainable logging?' and `Should we promote naked marriage?',
from Wikipedia.
For Hansard, we select debates related to the Middle-East War, the Russia--Ukraine War, and migrants \& refugees. From the source datasets, we annotate 140 randomly sampled \yc{potentially} emotional arguments\yc{\footnote{We use the emotion category classifier and emotional intensity predictor from \citet{Liu_2024} to retain arguments assigned an emotion label with high predicted intensity.}}
for two aspects: (1) whether they are target arguments, i.e., standalone claims for certain topics without any evidence, and (2) \ycr{which relevant contexts from the original documents are needed to interpret the corresponding situations.}
\yc{The annotation was done by one of the coauthors. We did not assess the annotation quality at this stage because it was used only to develop the data preprocessing tools; the quality of the final dataset was later evaluated end-to-end in \S\ref{sec:anno}.} 

Using these annotations, 
we then develop a prompt-based argument identifier, 
obtaining a Macro-F1 of 0.81 on the binary 
classification task, and a prompt-based context extractor, 
achieving a Macro-F1 of 0.59 based on exact span match, \yc{both with DeepSeek-V3.2 \citep{deepseekai2025deepseekv32pushingfrontieropen}}.\footnote{\yc{We evaluated the argument identifier on all 140 arguments and the context extractor on 84 contexts for 73 target arguments. Prompts were not systematically optimized but were iteratively revised by ChatGPT (\url{https://chatgpt.com/}) based on 5--10 inspected examples, so these results are diagnostic rather than strictly held-out. We tested the same prompts across several LLMs and report the best-performing ones, which were then used in the preprocessing.}}
Finally, we use these tools to select target arguments and extract their contexts from the original datasets.\footnote{We use the OpenRouter API (\url{https://openrouter.ai/}) for all LLMs in this work with default parameters.}   
\yc{The prompts used are presented in Appendices~\ref{app:prompt_claim} and~\ref{app:prompt_context}.}
We refer to Appendix~\ref{app:data} for more details.

\subsection{Synthetic Arguments \& Contexts}\label{sec:syn}

\begin{table}[!t]
\centering
\small
\setlength{\tabcolsep}{4pt}
\renewcommand{\arraystretch}{1.15}

\begin{tabularx}{\columnwidth}{
@{}
>{\raggedright\arraybackslash\hsize=1.2\hsize}X
>{\raggedright\arraybackslash\hsize=0.8\hsize}X
@{}
}
\toprule
\textbf{EA Injection} & \textbf{Context Perturbation} \\
\midrule

Claude Opus 4.6
\mbox{\citep{anthropic2026claudeopus46}}
    & Claude Opus 4.6 \\

DeepSeek-V3.2
\mbox{\citep{deepseekai2025deepseekv32pushingfrontieropen}}
    & DeepSeek-V3.2 \\

GPT-5.4
\mbox{\citep{openai2026gpt54}}
    & GPT-5.4 \\

Qwen3.6 Plus
\mbox{\citep{qwen2026qwen36plus}}
    & Qwen3.6 Plus \\

Gemini 3.1 Pro Preview
\mbox{\citep{googledeepmind2026gemini31pro}}
    & -- \\

MiniMax-M2.7
\mbox{\citep{minimax2026m27}}
    & -- \\

Mistral Small 2603
\mbox{\citep{mistral2026small4}}
    & -- \\

Qwen3.5-122B-A10B
\mbox{\citep{qwen2026qwen35}}
    & -- \\

\bottomrule
\end{tabularx}
\vspace{-.2cm}
\caption{Models used for EA injection and context
perturbation.}
\label{tab:models}
\vspace{-.4cm}
\end{table}

Here, we describe how we construct synthetic variants of the real-world argument--context pairs. We first apply an EA predictor to distinguish neutral arguments from emotional ones. We then apply an EA injector to synthesize emotional variants of the neutral arguments, and a context perturber to alter situation severity based on predefined reasons that may reduce \evi{} ratings.

\paragraph{Emotional Appeal Injection}
We prompt eight LLMs (see Table~\ref{tab:models}, left) to inject EAs into non-emotional arguments. We consider seven EA types---fear/threat, pity/compassion/sympathy, anger/indignation, hope, pride/flattery, guilt/shame, and loyalty/solidarity---and four intensity levels. The prompt, shown in Appendix~\ref{app:ea_prompt}, instructs models to first select three suitable EA types for a given topic and then generate arguments with EA levels from 2 (slight) to 5 (very strong) for each type. We encourage models to use emotional framing strategies, such as Storytelling and Loaded Words \citep{chen-etal-2026-emotionally}, and rhetorical figures, such as Repetition, Parallelism, and Rhetorical Questions \citep{10.1145/3744554}. We do not predefine the specific strategies or figures; models may use any strategy, figure, or combination thereof to achieve the target EA type and level.

\paragraph{Context Perturbation}
We prompt four frontier LLMs (see Table~\ref{tab:models}, right) to perturb the contexts given a topic and a target argument. As shown in Figure~\ref{fig:example}, the models first tag text spans related to the five factors that may affect \evi{}, and then perturb the tagged spans to change each factor separately while keeping the rest unchanged. If the original factor level is medium, we generate two variants: low and high. For example, if the original context cites a highly reliable source, such as \textit{a report from the United Nations}, the perturber may replace it with a less reliable source, such as \textit{an anonymous tweet}. The prompt is provided in Appendix~\ref{app:con_prompt}.
\ycr{We later confirm the reliability of this step in \S\ref{sec:anno}.}

\paragraph{Emotional Appeal Prediction}
We prompt GPT-5.4-mini to jointly predict the EA type and strength. The prompt is provided in Appendix~\ref{app:pre_ea_prompt}. We evaluate the predictor on 236 arguments annotated for EA type and strength: 76 are human-generated arguments from \S\ref{sec:data_real},\footnote{We include three arguments that contain evidence in addition to the 73 target arguments.} and 160 are synthetic arguments produced by EA injection into the non-emotional arguments from \S\ref{sec:data_real}. Details of the annotation procedure are provided in Appendix~\ref{app:emo_anno}.

Averaged over five evaluation runs, the predictor achieves a macro-F1 score of 0.72 for EA type classification and a Spearman correlation of 0.69 for EA strength rating. On human-generated arguments, it obtains a macro-F1 of 0.67 and a Spearman correlation of 0.56, compared with 0.77 and 0.69, respectively, on synthetic arguments. This trend is expected, as synthetic arguments typically make differences in EA type and strength more explicit and easier to distinguish.

\begin{table*}[!t]
\centering
\addtolength{\tabcolsep}{-0.12cm}
\resizebox{.9\textwidth}{!}{
\begin{tabular}{@{}llccccccc@{}}
\toprule
Source & Domain          & \# instances & \# topic & A-C groups                                                                & \makecell[l]{\# unique \\ arguments} & \makecell[l]{\# unique \\contexts}  & \makecell[l]{\# contexts\\ per argument} & \makecell[l]{\# arguments \\per context} \\ \midrule
Hansard     & \makecell[l]{Parliamentary \\debates} & 536                & 13       & \makecell[l]{R-R (134); R-S (134); \\S-R (134); S-S (134)} & 94 (58)                                              & 323 (97)                                         & 4.99                     & 1.45                     \\ \midrule
QAM         & Wikipedia            & 640                & 63       & \makecell[l]{R-R (160); R-S (160); \\ S-R (160); S-S (160)} & 130 (57)                                          & 418 (136)                                        & 4.46                     & 1.39                     \\ \midrule
Total       & -                  & 1176               & 76       & \makecell[l]{R-R (294); R-S (294); \\ S-R (294); S-S (294)} & 224  (115)                                         & 741 (233)                                          & 4.68                     & 1.42                    \\ \bottomrule
\end{tabular}}
\caption{Statistics of the annotation dataset. A-C groups indicate argument-context pair groups, where R denotes `real-world' and S `synthetic'; e.g., R-S denotes real world argument-synthetic context pairs. Values in brackets indicate the numbers of unique real-world (original) arguments/context. Distributions of EA types, EA strengths and context perturbations are provided in Table~\ref{tab:stats_2} (Appendix).}\label{tab:stats}
\vspace{-.4cm}
\end{table*}

\subsection{Human Annotation}\label{sec:anno}
\vspace{-.1cm}
\paragraph{Data}
We first apply the target argument identifier and context extractor to obtain argument--context pairs from the original datasets, and then use the EA predictor to label the EAs of the arguments. Non-emotional arguments undergo EA injection and are then labeled by the EA predictor, while contexts undergo context perturbation. We then sample argument--context pairs, balancing real-world and synthetic arguments and contexts, and pairing one argument with multiple contexts and one context with multiple arguments when possible.
The final dataset contains 1,176 argument--context pairs with their topics. Table~\ref{tab:stats} shows the final dataset statistics.

\yc{
To ensure \textbf{data quality}, we evaluate 100 random samples from the final dataset (50 Hansard and 50 QAM), covering 80 unique contexts, 64 unique arguments, and 20 instances per perturbation dimension. Two English-fluent student annotators assess whether each perturbation changes the intended dimension and whether the instance is valid and interpretable. The annotators achieve 75\% agreement and a Cohen's $\kappa$ of 0.57 on a 4-way classification task for evaluation (see Appendix~\ref{app:quality} %
for details). Averaged across annotators, 98.5\% of instances are judged valid, and 84.1\% match the intended perturbation direction. 
\ycr{We note that both annotator disagreements and mismatches between humans and LLMs largely reflect subjective threshold differences in judging whether a change is substantial enough to count as increased/decreased rather than unchanged. For example, one annotator considered both Congress and the American Medical Association credible institutions, while the other judged the latter to be more credible. However, overall, these results support the effectiveness of our preprocessing pipeline and context perturbations.}}

\yc{Since our analysis in \S\ref{sec:human} does not rely on the EA labels used to condition generation or those predicted by the EA predictor, we do not assess data quality along the EA dimensions.
} 
\ycr{}

\paragraph{Annotation}
We recruit \textbf{annotators} from the crowdsourcing platform Prolific\footnote{\url{https://www.prolific.com/}}. Using Prolific's pre-screening function, we select annotators who are native English speakers, based in the UK or US, hold a higher education degree, and have approval rates above 95\% with at least 100 submissions. Annotators are asked to evaluate the emotional justification of an argument given the topic and context, according to the evaluation framework introduced in \S\ref{sec:task}. 
We provide the annotation guidelines and screenshots of the annotation interface in Appendix~\ref{app:anno_detail}. 
We randomly split the 1176 annotation instances into  59 batches, each containing 20 items (except for one containing 16 items). Each batch is annotated by the same five annotators. In total, 295 unique annotators participate in the evaluation, costing around 2\se{.}5k Euros. 

\yc{To obtain more reliable consensus labels,} for each batch, we retain annotations only from the top three annotators, selected by their average quadratic weighted Cohen's $\kappa$ with the other annotators on the overall justification criterion. \yc{This reduces noise from annotators whose judgments were substantially less aligned with those of the rest of the batch.\footnote{\yc{We prioritize consensus rather than modeling annotator subjectivity, because reliable subjectivity analysis requires enough repeated judgments per annotator to distinguish individual tendencies from item-specific variation. However, emotion-related annotation is cognitively demanding; we therefore limit each crowdworker to 20 instances, which is insufficient for reliable subjectivity analysis.}}}

Table~\ref{tab:iaa} (Appendix) reports the average and maximum \textbf{annotation agreement} across batches among the top-3 annotators. We use Cohen's $\kappa$ and majority/full agreement for the single-label task (\eat), Jaccard similarity for the multi-label task (\evi{} Reduction Reasons), and quadratic weighted $\kappa$, Spearman's $\rho$, and majority/full agreement for ordinal ratings. All metrics except majority/full agreement are averaged over annotator pairs.
Overall, agreement is low, reflecting the subjectivity of the tasks, but is comparable to prior work. For example, \citet{chen-etal-2026-emotionally} report an average $\kappa$ of 0.30 for emotion category classification, compared with our 0.275 (8-way vs.\ 7-way). They also report Pearson's $r$ of 0.20 for EA level rating and 0.250 for convincingness, while our rating tasks obtain Spearman's $\rho$ of 0.285--0.390.\footnote{Pearson's $r$ and Spearman's $\rho$ measure different associations and are not directly comparable; this comparison is only a rough reference.}

We \textbf{aggregate labels} by majority vote. In cases where no majority is reached, we use the label from the annotator with the highest average agreement with the other annotators.

\paragraph{Label Distribution}
Label distributions are visualized in Appendix~\ref{app:label_dis}. On average, argument--context pairs receive \logic{} ratings of 1.5--1.7, indicating supportive relations. In both datasets, rating 3 
is the most frequent label (30\%--33\%), followed by ratings 2 (28\%) and 1 (17\%--18\%). 
\eat{}=None/unclear and \eas{}=1 are selected least often, suggesting that EA injection and sampling based on the EA predictor are generally successful.
For \eas{}, the most frequent labels are 4 and 3, with averages of 3.3 for QAM and 3.7 for Hansard. All other dimensions receive average ratings of 3.4--3.8, with 4 as the most frequent label, suggesting that the EAs in our data are mostly perceived as justified, consistently across criteria. Hansard consistently receives higher scores than QAM, especially for \norm{} (3.8 vs.\ 3.5). For \evi{} reduction reasons, \emph{no reduction / fully justified} is selected most often, while spatial and temporal proximity are selected least often.

\emph{We exclude items with majority labels of \eat{}=None/unclear or \eas{}=1 from further analysis, retaining 1,028 items.}

\section{Human Perceived Justification of Emotional Appeal}
\label{sec:human}

In this section, we examine whether human-perceived EA justification aligns with the literature.

\begin{figure}[!ht]
    \centering
    \includegraphics[width=\linewidth]{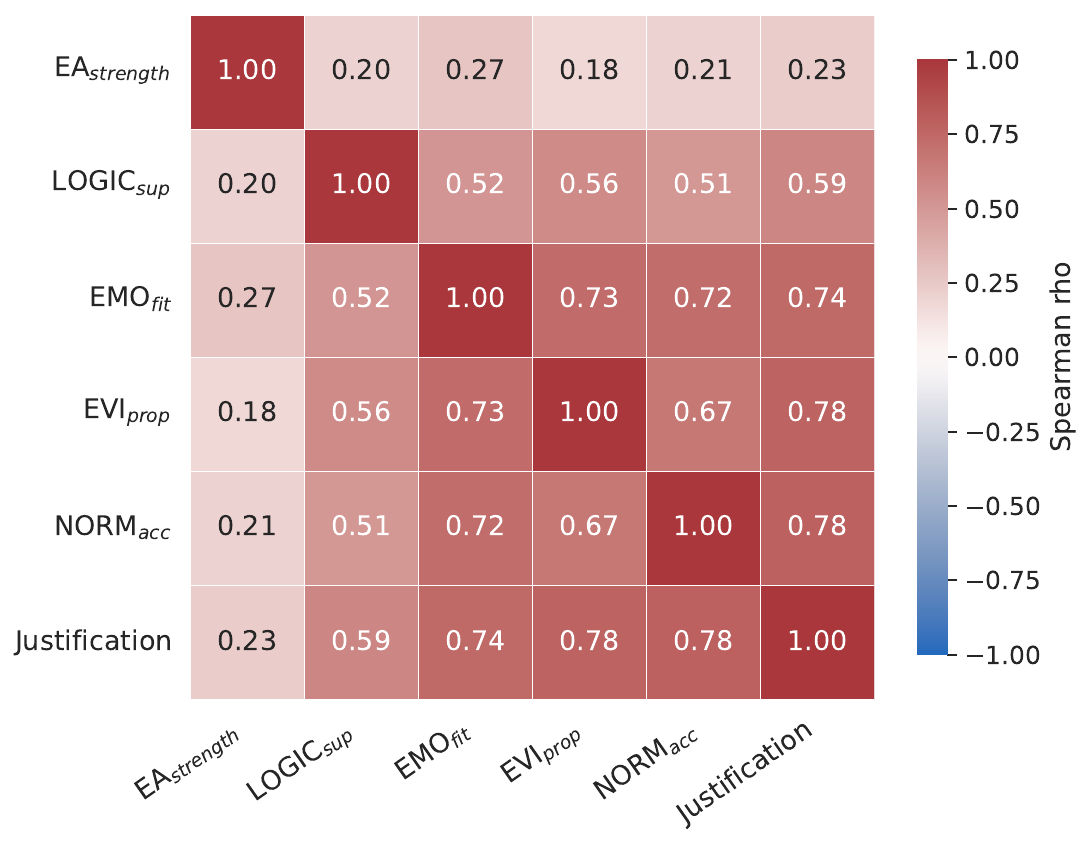}
    \vspace{-.8cm}
    \caption{Spearman correlation between annotated criteria.}
    \label{fig:spearman}
    \vspace{-.6cm}
\end{figure}

\begin{table*}[!t]
\centering
\addtolength{\tabcolsep}{-0.12cm}
\resizebox{\textwidth}{!}{
\begin{tabular}{@{}lcccc|cccc|cccc@{}}
\toprule
        & \multicolumn{4}{c|}{Hansard}                  & \multicolumn{4}{c|}{QAM}                      & \multicolumn{4}{c}{ALL}                      \\ \midrule
EA & \logic{} & \fit{} & \evi{} & \norm{} & \logic{} & \fit{} & \evi{} & \norm{} & \logic{} & \fit{} & \evi{} & \norm{} \\ \midrule
pity    & 0.184   & 0.230     & \underline{0.294}      & \textbf{0.361}      & 0.158   &\underline{0.237}    & 0.149*     & \textbf{0.428}      & 0.168   & 0.210     & \underline{0.246}      & \textbf{0.376}      \\
anger   & 0.053*  & 0.129*   & 0.048*     & \textbf{0.727}      & 0.072*  & 0.143*   & \underline{0.299}      & \textbf{0.445}      & 0.067   & \underline{0.141}    & 0.137      & \textbf{0.602}      \\
guilt   & 0.141*  & \underline{0.402}    & \textbf{0.448}      & 0.090*     & \underline{0.187}   & 0.185*   & \textbf{0.588}      & -0.046*    & 0.179   & \underline{0.268}    & \textbf{0.502}      & 0.004*     \\
hope    & 0.031*  & 0.132*   & \textbf{0.470}       & \underline{0.245}      & 0.055*  & -0.025*  & 0.156*     & \textbf{0.618}      & 0.062*  & -0.028*  & \underline{0.347}      & \textbf{0.472}      \\
fear    & 0.107   & \underline{0.263}    & \textbf{0.361}      & 0.212      & 0.044*  & 0.120*   & \underline{0.282}      & \textbf{0.528}      & 0.072   & 0.185    & \underline{0.331}      & \textbf{0.387}      \\
loyalty & 0.045*  & 0.225    & \textbf{0.428}      & \underline{0.279}      & -0.356* & 0.504*   & -0.173*    & 0.589*     & 0.052*  & \underline{0.312}    & \textbf{0.409}      & 0.258      \\
pride   & -1.842* & -0.442*  & 3.958*     & -0.116*    & -0.055* & 0.341*   & -0.012*    & 0.752*     & -0.011* & 0.121*   & 0.117*     & 0.653*     \\ \midrule
ALL     & 0.091   & 0.233    & \underline{0.307}      & \textbf{0.365}      & 0.092   & 0.116    & \underline{0.270}      & \textbf{0.459}      & 0.097   & 0.165    & \underline{0.301}     & \textbf{0.399}      \\ 
\bottomrule
\end{tabular}}
\caption{Coefficients of each evaluation criterion in the mixed-effects regression model predicting justification scores. Values marked with * are statistically \textbf{in}significant. The highest significant coefficient is bolded, and the second highest significant coefficient is underlined.}\label{tab:mixed}
\vspace{-.6cm}
\end{table*}

\paragraph{Justification criteria rise and fall together.}
Figure~\ref{fig:spearman} shows Spearman correlations among the ordinal criteria. All sub-criteria correlate positively with EA justification, with moderate to strong correlations ($0.59$--$0.78$). \evi{} and \norm{} show the strongest correlations, while \logic{} is the weakest. The sub-criteria also correlate with each other ($0.52$--$0.73$), suggesting that they reflect a shared latent judgment space rather than independent evaluation axes.

We further apply principal component analysis (PCA) to the annotation scores separately for Hansard and QAM. As shown in Figure~\ref{fig:pca} (Appendix), PC1 explains most variance in both domains (71.5\%--72.6\%), confirming that the criteria largely move together. PC2 (11.5\%--13.1\%) consistently contrasts \logic{} with the emotion-related criteria, highlighting the distinction between logical support and emotion-based justification. Overall, human judgments are dominated by a shared justification factor, with smaller components capturing finer-grained trade-offs among the sub-criteria.

\paragraph{Evidential Proportionality and Normative Acceptability matter most.}

To examine which criteria explain overall EA justification, we fit mixed-effects regression models predicting EA justification from the four evaluation criteria. The results should be interpreted as conditional associations: they indicate which criteria distinguish more justified from less justified EAs \yc{in our data}, given the observed annotation scores. As Table~\ref{tab:mixed} %
shows: Overall, EA justification is most strongly associated with \norm{} and \evi{}. On all data, \norm{} has the largest coefficient ($\beta=.399$), followed by \evi{} ($\beta=.301$), while \logic{} has the smallest coefficient ($\beta=.097$). This suggests that annotators judge EAs as more justified when they are socially and culturally acceptable, and when their strength is proportionate to the severity of the situation evidenced by the context. 
\yc{The smaller coefficients for \logic{} and \fit{} may arise for different reasons: \logic{} varies little because most contexts already support the argument, whereas \fit{} shows variation comparable to \evi{} and \norm{} but contributes less uniquely to overall justification once the other criteria are considered.}

The domain difference may reflect that EAs are less expected in QAM than in Hansard. QAM is more informational and less rhetorical, so annotators may rely more on whether an EA is normatively acceptable in that context. Accordingly, \norm{} is the strongest predictor in QAM ($\beta=.459$), while \fit{} is weaker ($\beta=.116$). By contrast, because EAs are more expected in parliamentary debates, Hansard judgments are less dominated by \norm{} and are more evenly associated with \fit{}, \evi{}, and \norm{} ($\beta=0.233$--$0.365$).

Emotion-specific results further suggest that different EAs are justified through different criteria. Pity and anger are mainly associated with \norm{}, indicating that they are judged more justified when perceived as socially more appropriate. %
Guilt differs: \evi{} is the strongest and most consistent predictor across Hansard ($\beta=.448$), QAM ($\beta=.588$), and the combined data ($\beta=.502$), while \norm{} is insignificant in all three models. This suggests that guilt appeals are justified mainly when the premise provides sufficient evidence for assigning responsibility or blame. 

\ycr{
Tables~\ref{tab:ex_qam} and ~\ref{tab:ex_hansard} in the appendix provide examples identified as having high and low EA justification in our annotated dataset.
}

\paragraph{EAs can remain partially justified without \yc{logical} support.}

Figure~\ref{fig:spearman} shows a moderate positive correlation between \logic{} and EA justification (0.59), indicating that higher \yc{logical} support is associated with higher perceived justification. 
We further compare justification scores across opposing (\logic{}$<0$), neutral (\logic{}$=0$), and supporting (\logic{}$>0$) contexts in Figure~\ref{fig:bar_pie} (left). As expected, justification is lowest in opposing contexts and highest in supporting contexts. However, non-supporting contexts still receive mean/median scores of 2.42--2.72 and 2--3, i.e., between slightly and partly justified.
This suggests that lack of \yc{logical} support reduces perceived justification, but does not necessarily make an EA completely unjustified. This complicates Def.~1: annotators may still assign partial justification when other criteria, such as \fit{}, \evi{}, or \norm{}, are satisfied, \yc{even if the premise does not justify the conclusion.}

\paragraph{Strong EAs are not necessarily unjustified.}
Figure~\ref{fig:spearman} shows a weak positive correlation between EA strength and EA justification ($\rho=.23$), suggesting that stronger EAs are not generally judged as less justified. However, this may be confounded by context: more severe situations may license both stronger EAs and higher justification scores. We therefore compare 893 argument pairs with the same context, where one argument has a stronger EA than the other. As shown in Figure~\ref{fig:bar_pie} (right) in the appendix, only 20.7\% of pairs show the stronger EA receiving a lower justification score.
Thus, even when context is fixed, we find no evidence that stronger EAs are perceived as less justified. 
\ycr{This is consistent with}
\evi{}: what matters is not EA strength per se, but whether the strength is proportionate to the situation evidenced by the context.

\begin{figure}[!ht]
    \centering
    \includegraphics[width=\linewidth]{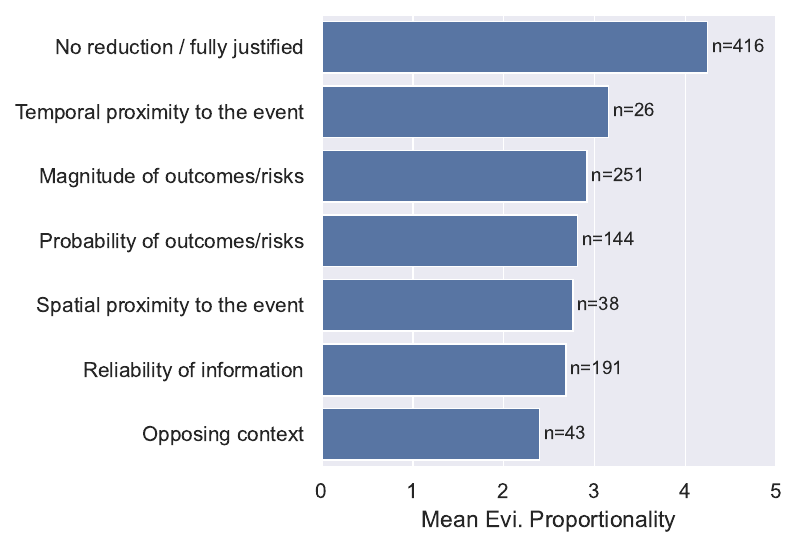}
    \vspace{-.6cm}
    \caption{Average \evi{} by reason category for reductions in evidential proportionality.}
    \label{fig:reason}
    \vspace{-.6cm}
\end{figure}

\paragraph{Source reliability most strongly reduces evidential proportionality.}

We examine average \evi{} ratings for each reduction reason. For multi-label annotations, an item is counted under a category if that category appears among its labels. Figure~\ref{fig:reason} %
shows the frequency of each reason and its average \evi{} rating. As expected, \textit{No reduction / fully justified} receives the highest ratings, while \textit{Opposing context} receives the lowest.

Among the predefined factors, \textit{Reliability of information source} receives the lowest \evi{} ratings ($2.69$ vs.\ $2.76$--$3.16$). Its ratings are significantly lower than those for \textit{Temporal proximity} ($3.16$; $p=.047$) and \textit{Magnitude of outcomes/risks} ($2.92$; $p=.018$). This suggests that unreliable sources may undermine the evidential basis of the premise itself: when the information is not trusted, the EA may be seen as poorly warranted regardless of the premise content.

\section{Can LLMs Predict the Justification of Emotional Appeals?}\label{sec:llm_predict}

We conduct a small-scale experiment on EA justification prediction with two LLMs, DeepSeek-V3.2 and GPT-5.4-mini, which perform well on EA prediction and context extraction in our previous evaluation (\S\ref{sec:data}). We use automatic prompt optimization: a refiner model iteratively analyzes prediction errors and revises the prompt for the predictor model; the refiner and predictor use the same model.
\yc{After bootstrapping the initial prompts from the gold examples, at each iteration, the refiner receives summaries of previous rounds and current correct and incorrect predictions. To optimize Spearman correlation, we additionally provide highest-severity ranking inversions and correctly ranked pairs. The refiner then revises the prompt while preserving correct behavior. See Appendix~\ref{app:prompt_eaj} for the refiner prompts.}

We optimize prompts on a larger portion of the data and evaluate them on a held-out test set. To avoid near-duplicate leakage, we group items sharing the same original argument or context, and randomly split groups into optimization and test sets at an 8:2 ratio. We also preserve the Hansard/QAM proportions in the test set. We repeat this process over three random splits and report the average Spearman correlations between model predictions and human judgments in Table~\ref{tab:llm_pred}.

\begin{table}[!t]
\centering
\resizebox{.8\columnwidth}{!}{
\begin{tabular}{@{}lcc@{}}
\toprule
  & GPT-5.4-mini & DeepSeek-V3.2 \\ \midrule
\logic{} & 0.363           & 0.400           \\
\fit{} & 0.351           & 0.336           \\
\evi & 0.327           & 0.367           \\
\norm{} & 0.135           & 0.370           \\
overall & 0.312           & 0.350           \\ \bottomrule
\end{tabular}}
\vspace{-.2cm}
\caption{Spearman correlations between model predictions and human judgments of EA justification.}\label{tab:llm_pred}
\vspace{-.6cm}
\end{table}

Overall, correlations are low, indicating the difficulty of the task. Both models perform best on \logic{} prediction: GPT-5.4-mini obtains $0.363$ on \logic{} compared with $0.135$--$0.351$ on other criteria, while DeepSeek-V3.2 obtains $0.400$ compared with $0.336$--$0.370$. This is expected, as \logic{} captures a more objective relation between context and argument. In contrast, emotion-related criteria are harder to predict, especially \norm{} for GPT-5.4-mini ($0.135$). These results suggest that simple prompting 
is insufficient for EA justification prediction.

\section{Conclusion}

In this work, we introduce the task of predicting the justification of EAs in arguments and collect the first human-annotated dataset for EA justification. We ground the annotation in an evaluation framework built on philosophical literature. We show that the justification criteria share a latent judgment space rather than forming independent evaluation axes. Among these criteria, \evi{} and \norm{} are most strongly associated with overall justification. We also find that an EA without logical support is not always perceived as fully unjustified, and that stronger EAs are not necessarily perceived as less justified, partly complicating existing theories. Finally, we show that predicting EA justification remains challenging for LLMs, especially for emotion-related criteria.

\yc{Our current task evaluates the perceived EA justification conditional on the provided context and does not assess its factuality or completeness.} Future work can build EA justification predictors for real-world scenarios by \yc{retrieving sufficiently complete context and incorporating fact-checking to verify its factuality}. Such predictors can also support argument generation with EAs while avoiding fallacious persuasion.

\section{Limitations \& Ethical Concerns}

This work has several \textbf{limitations}. (1) Our task formulation simplifies full argument reconstruction by evaluating an argument with respect to a paired context and topic. This makes annotation feasible, but does not cover more complex argumentative structures where EAs interact with multiple premises, implicit assumptions, or longer discourse contexts.
(2) EA justification is inherently subjective. Although our agreement is comparable to related emotion annotation tasks, annotators may still differ in their judgments, especially for emotion-related criteria such as \norm{}. Our labels should therefore be interpreted as human-perceived justification rather than fully objective truth.
(3) Our data is limited in domain and language. We focus on English arguments from Hansard and QAM, with annotators based in the UK or US. Since \norm{} depends on social and cultural expectations, the framework may not directly generalize to other languages, cultures, or discourse communities.
(4) Part of our dataset is synthetically generated. Although synthetic variants support controlled analysis, they may contain artifacts from LLM generation or context perturbation. \yc{(5)} 
Our preprocessing tools, including argument identification, context extraction, and EA prediction, are imperfect and may introduce noise.  \yc{(6)} 
Our model experiments are small-scale and prompt-based, covering only two LLMs; stronger supervised or retrieval-augmented models may perform differently.

\paragraph{Ethical Concerns}
This work studies EAs because they can influence persuasion and may be used for manipulation, especially when generated at scale by LLMs. A potential risk is that our framework or dataset could be misused to generate more persuasive emotional arguments. To reduce this risk, we frame the task as evaluating whether an EA is justified in context, with the goal of distinguishing warranted emotional persuasion from unjustified or manipulative uses.

Our data includes political and socially sensitive topics, such as war, migration, and refugees. These topics may evoke strong emotions and may reflect biases in the original sources, LLM-generated variants, or annotator judgments. In particular, \norm{} is culturally dependent, so the resulting labels should not be treated as universal judgments of acceptability.

We recruited annotators through Prolific and asked them to evaluate arguments according to explicit guidelines. Since the task involves subjective and potentially sensitive judgments, we aggregate multiple annotations and analyze disagreement rather than assuming a single objective label. 

\paragraph{AI Assistance}
We used AI tools for text refinement and to assist with experiments, such as preparing crowdsourcing tasks and partial coding. 

\paragraph{Annotator Privacy and Consent}
We obtained annotators' consent through Prolific when collecting their annotations. We will not release any data containing annotators' personal information. All released annotations will be anonymized.

\section*{Acknowledgements}
We thank the anonymous reviewers for their thoughtful feedback, which has helped strengthen the paper. The NLLG Lab gratefully acknowledges support from the German Research Foundation (DFG) via the Heisenberg Grant EG 375/5-1. This project has been conducted as part of the EMCONA (The Interplay of Emotions and Convincingness in Arguments) project, which is funded
by the DFG (project EG-375/9-1).

\bibliography{custom}


\appendix

\section{Dataset Creation}\label{app:data}
\paragraph{Data Source}
We first obtain noisy argument-context pairs and their topics from \textbf{two data sources}:
(1) We crawl \textbf{the UK parliamentary debates (Hansard)} for year 2025 via the official API;\footnote{https://hansard-api.parliament.uk} the covered period is 2025/01/06--2025/12/18. We then filter debates based on title keywords. Specifically, we retain debates whose titles contain keywords related to our three focal topics: the Middle-East War, the Russia-Ukraine War, and Migrants \& Refugees. The predefined keyword lists for each topic are provided in Table~\ref{tab:hansard-keywords}. In addition, we exclude debates whose titles mention Bills, Readings, or Amendments, as these debates typically require detailed background knowledge of the referenced legislative materials. 209 candidate debates were retained after this step.


\begin{table}[h]
\centering
\small
\begin{tabularx}{\linewidth}{@{}lX@{}}
\toprule
\textbf{Topic} & \textbf{Keywords} \\
\midrule
Russia--Ukraine
& \texttt{russia}, \texttt{ukraine} \\ \midrule

Middle East
& \texttt{israel}, \texttt{palestine}, \texttt{gaza},
  \texttt{hamas}, \texttt{hezbollah}, \texttt{hizbollah},
  \texttt{hisbollah}, \texttt{iran} \\ \midrule

Migrants \& Refugees
& \texttt{asylum}, \texttt{deportation}, \texttt{expulsion},
  \texttt{displacement}, \texttt{migration}, \texttt{migrant},
  \texttt{immigrant}, \texttt{refugee}, \texttt{repatriation},
  \texttt{integrat}, \texttt{invasion} \\
\bottomrule
\end{tabularx}
\caption{Keywords used to select Hansard debates through
case-insensitive substring matching on debate titles. Debates with titles
containing \emph{Bill}, \emph{Reading}, or \emph{Amendment} were excluded.}
\label{tab:hansard-keywords}
\end{table}

To get candidate argument-context pairs, we apply heuristic prompt-based end-to-end argument mining pipeline to get the main topic, the claims and the contexts relevant to the claims, and the clusters of similar claims, for each debate. To find frequent shared claims across debates, we then use k-means to cluster the claim clusters. We then pair each claim and each context belonging to the same cluster to get the candidate pairs. 

(2) \textbf{QAM} \citep{guo-etal-2023-aqe} consists of argument quadruplets for a given topic, each containing a claim sentence, the stance of claim towards the topic, an evidence sentence supporting the claim, and the evidence type. extracted from Wikipedia. Evidence types include Research, Case, Expert, Explanation, and Other; we refer readers to the original paper for their definitions.
We then retain the claim-evidence pairs with evidence types of expert and research, because these two types of evidence connect to factual evidence like statistics from sources like authorities, scholars or research, which can provide information for understanding the situation.

\paragraph{Preprocessing}
We first obtain noisy candidate argument--context pairs from real-world argument datasets with human or LLM labels for manual verification. Since non-emotional arguments tend to dominate argument datasets, we use prior model predictions to guide sampling and (potentially) obtain a more balanced annotation set across emotional appeal types and strength. Specifically, we apply the emotion category classifier and the emotional intensity predictor from \citet{Liu_2024} to the arguments,\footnote{We use EmoT5 for emotion category classification and EmoLLaMA-chat-13B for emotional intensity prediction, as these models achieve the best performance on the respective tasks in the original paper.} and sample 140 emotional arguments, based on the resulting prior labels, and their contexts for manual verification.

73 \textbf{arguments} are identified as target arguments. A target argument must be a standalone claim without any evidence. We further require it to be either a direct proposition on the topic or a top-level argument that expresses a clear stance toward the topic. Lower-level argumentative content that supports or explains the target argument is then treated as its context. For example, for the topic \textit{whether smoking should be banned}, a direct proposition can be \textit{Smoking should be banned}. A top-level argument can be \textit{Smoking is bad for health}, which directly supports the proposition in an argument such as \textit{Smoking should be banned because it is bad for health}. A lower-level argument, such as \textit{Smokers are more likely to get cancer than non-smokers}, is then treated as part of the context for either the direct proposition or the top-level argument, since it provides more concrete information for evaluating whether the target argument is justified. Such lower-level arguments are often connected to factual evidence, e.g., \textit{Recent research found that 6\% of smokers develop cancer, compared to 1\% of non-smokers, implying that smokers are more likely to get cancer than non-smokers}.

As for \textbf{contexts}, we retrieve the original documents containing the candidate contexts and re-extract the contexts for each target argument. In total, we extract 84 contexts for the 73 target arguments. We require each context to provide concrete information that can be used to evaluate the situation. For example, for the target argument \textit{Smoking is bad for health}, \textit{Recent research found that 6\% of smokers develop cancer, compared to 1\% of non-smokers} is a valid context, whereas \textit{Smoking harms nearly every organ of the body} is not, because it states a factual claim without providing evidence that can be evaluated. 
We further require each context to be a minimal contiguous text span that provides a complete picture of the relevant information. For example, \textit{Recent research found that 6\% of smokers develop cancer, compared to 1\% of non-smokers. This implies that smokers are more likely to get cancer than non-smokers} should be extracted as a single context rather than two separate contexts, since the two sentences together provide the relevant evidence and its implication.

\section{Prompts}\label{app:prompt}

\subsection{Claim Identifier}\label{app:prompt_claim}
\begin{lstlisting}[style=promptstyle]
system prompt:
You are an expert annotation assistant for claim identification in argument mining.

Your task is to decide whether a candidate claim should be selected for a given topic.

Selection definition:

- Select the claim if it is a standalone, direct proposition expressing a clear stance toward the topic.

- Also select the claim if it is a broad, highest-level argument that directly supports or opposes the topic without relying on a formal citation, quotation, named source, concrete case, statistic, anecdote, or detailed evidence.

- General self-contained reasons about harms, benefits, risks, social acceptability, resource scarcity, morality, practicality, character/fitness of relevant actors, functional usefulness of the topic object/action, or comparative outcomes can be selected when they function as top-level reasons for or against the topic's same main action/object/outcome.

- For ``Should/Shouldn't'' questions, infer the proposed action and be inclusive of broad generalized reasons bearing directly on that action or its object.

- A claim may begin with a discourse marker or an elliptical reason marker such as ``because'', ``when'', ``especially when'', ``hence'', or ``but''; select it if the remaining content states a clear generalized reason.

- Light generic attribution is not disqualifying when the candidate still states a complete general proposition.

- Complete claims with simple pronouns such as ``they'' or ``these people'' may be selected when the pronoun clearly refers to the topic's main actor/object.

- For declarative or policy-proposition topics, require the claim to address the same main object, action, and outcome as the topic.

- Ignore punctuation, case, minor discourse markers, outline numbering, and bracketed cleanup tags when judging the claim.

Do not select claims that are only evidence, examples, statistics, study findings, citations, quotations, anecdotes, concrete cases, background facts, definitions, implementation details, or lower-level support for another argument.

Do not select a claim merely because it is topically related.

Do not use the topic to complete a missing subject, verb, or object in the claim.

Reply with exactly one lowercase word: \texttt{true} or \texttt{false}.

Do not use JSON. Do not include an explanation.

user prompt:
Topic:

{topic}

Claim:

{claim}

Should this claim be selected for the topic?

Decision checklist:

- 1. Normalize mentally for punctuation, case, minor discourse markers, outline numbering, and bracketed cleanup tags.

- 2. Identify the topic form.

- If it is a ``Should/Shouldn't'' question, infer the action and allow broad generalized top-level reasons for or against that same action or its object.

- If it is a declarative/policy proposition, require a close match to the same main object, action, and outcome, or a clean top-level reason specifically tied to that proposition.

- 3. Answer true if the claim itself is a standalone proposition that directly takes a stance toward the topic, or states a broad highest-level reason directly for or against it.

- 4. Treat generalized harms, benefits, risks, moral/practical reasons, actor-fitness reasons, social acceptability claims, resource-scarcity claims, functional-usefulness claims, and comparative outcome claims as selectable when they are self-contained.

- 5. Do not reject merely for hedging, imperfect grammar, light generic attribution, comparative prediction, conditional wording, or an elliptical reason marker.

- 6. Generic attribution may be ignored when no person/organization is named and the remaining content is a general top-level reason.

- 7. A complete proposition with a simple pronoun subject may be true if the pronoun clearly refers to the topic's main actor/object.

- 8. Answer false if the claim is mainly a citation, quotation, named-source statement, statistic, study result, anecdote, concrete example, evidential bundle, background/current-condition fact, implementation detail, or lower-level support.

- 9. Answer false if the claim is only related, shifts to a different object/action/location, states an adjacent principle, gives only examples, is a vague call to act, or relies on unclear deictic words.

- 10. If removing attribution, citation, quotation, evidence, concrete details, or surrounding context leaves only a fragment or a different/vaguer proposition than the topic, answer false.

- 11. Reply with exactly one lowercase word: true or false. Do not use JSON.
\end{lstlisting}

\subsection{Context Extractor}\label{app:prompt_context}

\begin{lstlisting}[style=promptstyle]
System Prompt:
You are an expert annotation assistant for extracting target context spans.

Your task is to extract the target context spans from a source text for a given claim.

A target context is narrower than generic support, related background, or any sentence that could help assess the claim. It is the direct factual, provenance, example, statistic, event, legal/factual finding, concrete action, or setup anchor that the claim depends on or points to.

Extraction rules:
- Extract only text that appears verbatim in the source text.
- Every extracted context must be one exact contiguous substring of the source text.
- Do not paraphrase, summarize, rewrite, change casing, normalize punctuation, or stitch together non-contiguous text.
- Preserve the source text's capitalization, punctuation, quotation style, and whitespace as much as possible.
- Prefer complete sentences and complete multi-sentence evidence units. Use sentence fragments only when trimming a header, salutation, rhetorical lead-in, claim sentence, or detachable claim-restatement phrase is necessary.
- Return only valid JSON in the schema {"contexts": ["..."]}. If no direct target context exists, return {"contexts": []}.

What counts as target context:
- named sources, authorities, reports, studies, statistics, citations, or attributed factual statements
- concrete events, examples, cases, numbers, risks, outcomes, dates, locations, legal findings, or factual background that directly grounds the claim
- concrete support actions, funding, training, aid delivery, military support details, bilateral donor commitments, delivery mechanisms, or authoritative quotations when the claim is a broad support/solidarity/continuing-support claim
- immediate explanatory setup needed to understand a concrete example, study, statistic, finding, or action

What does not count:
- the claim itself, a verbatim repetition, or a close restatement of the same proposition
- neighboring claims, conclusions, recommendations, policy demands, ministerial questions, promises, slogans, predictions, final appeals, moral interpretations, rhetorical intensification, or same-side argumentative elaboration
- broad topic background that does not directly match the claim's central subject and predicate

Hard precision rules:
- If a sentence contains the exact claim or a close restatement as its main clause or as one coordinate clause, exclude the whole sentence. Do not extract any other clause, list item, phrase, or trailing material from that same sentence, even if the extracted substring would omit the claim wording.
- If a selected factual sentence ends with a detachable appositive or concluding phrase that merely restates the claim, such as ", which shows that ...", ", a clear commitment that ...", "must be condemned", or "this proves...", trim off that phrase and keep only the preceding exact factual substring.
- Trim non-evidentiary in-sentence lead-ins such as "It is probably a good time to acknowledge that" when a clean concrete factual clause remains. Do not trim real source/authority attribution such as "Researchers found", "As Justice X noted", or "the report concluded".
- Match the strength and predicate of the claim. For strong claims such as "open warfare", "annihilation", "genocide", "ethnic cleansing", "eradication", trafficking, or systematic unlawful conduct, cautious concern language, "reports of" language, unsupported speaker recollections, or general worry is not enough unless it contains concrete scale/casualty facts or an authoritative legal/factual finding matching that predicate.
- For institutional recommendations about child trafficking, forced transfer, intelligence sharing, or support efforts, prefer aggregate returned/missing-child numbers, authoritative legal breaches, and systematic child harm/casualty facts. Reject first-person visit narratives, meeting setup, tribute sentences, "we heard how..." anecdotal chains, and individual stories unless the claim specifically depends on that individual case.
- For aid-access or humanitarian-supply claims, prefer concrete dependency, threat-to-life, deprivation, or humanitarian harm facts over standalone illegality/condemnation sentences.
- Reject unsupported predictions such as "numbers are likely to rise" unless the claim itself is about that prediction and the source/finding directly supports it.

Evidence-unit boundary rules:
- Construct evidence units before producing output. Do not output individual sentences and then leave adjacent pieces split.
- If adjacent sentences jointly form one target context unit, return that whole contiguous unit as one context instead of splitting it.
- Same-unit patterns include: source-plus-finding, study-description-plus-method-plus-result, statistic-plus-scope, assessment-plus-statistic, setup-plus-result, comparison-plus-interpretation, event-plus-impact, attack-plus-scale-plus-impact, legal finding-plus-implication, concrete action-plus-continuation, observed-case setup-plus-case, problem-plus-response, aid/funding package, training/support package, short historical/alliance point, numerical comparison sequence, and one concrete example.
- If an evidentiary assessment immediately introduces numbers, figures, costs, harms, or scale and the next sentence gives the concrete statistic or direct impact, keep the contiguous assessment-plus-statistic/impact unit together.
- Merge adjacent numerical/statistical sentences when later sentences calculate net figures, contrast categories, give proportions, or complete the same comparison needed for the claim.
- Merge adjacent cumulative casualty/harm sentences when they jointly establish scale or systematic harm. Do not output component sentences separately.
- Do not merge a concrete harm/casualty/statistic unit with a separate standalone legal conclusion, broad moral characterization, or separate policy response merely because it is adjacent. In particular, keep a sentence such as "Russia's actions violate..." separate from the preceding casualty/systematic-harm unit unless the claim specifically requires one combined legal-characterization unit.
- If adjacent sentences are independent support actions using different instruments, actors, recipients, or programmes, return them as separate contexts unless they are explicitly signaled as one package. For example, partner-delivered funding and British Army training are separate units.
- For an amount/commitment sentence, stop at the end of that concrete commitment unless the following sentence adds another concrete amount, named programme, recipient, delivery mechanism, or signaled package detail. Do not append broad generic follow-up such as "we are supporting efforts..." or "we are helping..." without concrete details.
- If a harm/problem sentence is immediately followed by "That is why", "That includes", "This is why", or similar wording plus concrete aid, repairs, funding, training, or humanitarian support tied to the same problem, merge the whole harm-plus-response chain into one contiguous context, trimming any trailing claim-restatement phrase if needed.
- If separate non-adjacent parts independently provide direct target context anchors, return them as separate contexts.
- If two candidate contexts overlap or one contains the other, keep the most useful self-contained span and avoid duplicate nested extractions.
- Before final output, if selected contexts are adjacent sentences from the same evidence unit, replace them with one merged contiguous span; if one span overgrew across independent units, split it.

Claim-type guidance:
- For recommendation, investigation, or policy-action claims, the target context may be the concrete harm, statistic, legal breach, source finding, or factual problem motivating the recommendation. Do not extract the recommendation, question, appeal, or general policy option itself.
- If a policy-action claim includes a specific mechanism or outcome, prefer evidence for that exact mechanism/outcome over general endorsements, source opinions, integration/public-finance arguments, pull-factor rebuttals, or broad policy background.
- For problem/ineffectiveness claims, later concrete trend, volume, cost, capacity, or named-source statistics may be target contexts even if they do not repeat the exact wording. Prefer independent later statistics over clauses from the claim-containing sentence.
- For broad stance, solidarity, or continuing-support claims, scan the whole relevant passage for concrete attack/harm anchors, humanitarian effects, responder attacks, demographic support, partner-delivered funding, military support/training, bilateral donor commitments, named aid packages, and authoritative quoted commitments. Include concrete support actions even if they do not repeat the stance wording.
- For broad support claims, a sentence saying a named minister or official "set out details of our military support" may count when it directly states continuing work with the supported partner to provide what they need.
- For authoritative quotation units, include the attribution with the quotation. If consecutive sentences in the same conclusion paragraph quote or paraphrase the same authority's peace principle, current attack assessment, and Ukraine's resilience, keep that attributed quote/assessment unit together when it directly supports the broad stance.
- In QAM/pro-con/article text, be conservative with headings and generic explanatory prose. However, when the claim is about concrete social effects, stress, pressure, harms, or observed practices, consecutive factual examples under the same question/answer are observed-case units and may be target context even without a named study.
- For QAM/article observed cases, merge the full consecutive setup/scope/case sequence, including intermediate sentences such as general local practice followed by a specific reported case. Do not split the first behavior sentence from a later "there have been cases..." sentence if they are part of the same observed example sequence.
- For research/study claims, if the claim depends on a research result, output the contiguous research unit needed for that result. If the claim only needs provenance or category information, a citation/source fragment may be sufficient; do not automatically include downstream mechanism/background.
- For continuing-action claims, include adjacent sentences that identify the concrete ongoing effort/status and its support, but not the claim sentence itself.

Special text handling:
- Do not return titles, section headers, topic labels, speaker salutations, or opening claim sentences as context. Trim them away if a clean factual substring remains.
- Do not add a preceding organization/source-meeting/setup sentence when the following sentence contains the concrete result, number, finding, or harm and is understandable enough as evidence, even if it uses a pronoun such as "it" or "they".
- If a relevant sentence is followed by a short acknowledgement or transition that is grammatically or paragraph-contiguously part of the same concrete support unit and still names the affected group, keep it with that unit rather than trimming to a fragmentary boundary.

User Prompt:
Topic:
{topic}

Extract the target context spans from the source text below.

Claim:
{claim}

Document:
{document}

Selection checklist:
1. Identify the claim's central subject and predicate; ignore spans tied only to the broader topic or a different sub-issue.
2. Classify the claim type:
   - specific factual/evaluative claim,
   - recommendation, investigation, or policy-action claim,
   - problem/ineffectiveness claim,
   - broad stance/solidarity/continuing-support claim,
   - QAM/pro-con/article heading, value claim, or observed-case claim.
3. Identify candidate evidence units, not just sentences. A unit may be one sentence, a clause trimmed from a rhetorical lead-in, or a contiguous run of adjacent sentences.
4. Remove the claim itself and any sentence containing the claim/restatement as a main or coordinate clause. Do not salvage sibling clauses, list items, or phrases from that same sentence.
5. Remove candidates that are only emotional appeal, value framing, generic explanation, hypothetical reasoning, prediction, recommendation, policy demand, question, final appeal, promise, speaker recollection/opinion without external anchor, or tangential detail.
6. For strong labels such as genocide, annihilation, ethnic cleansing, trafficking, or unlawful conduct, require concrete scale/casualty facts or an authoritative legal/factual finding; do not use mild concern, reports, or the speaker's own allegation alone.
7. For recommendation/policy-action claims, keep concrete motivating facts, legal/statistical harms, or evidence for the exact mechanism/outcome; reject general endorsements, anecdotes, meeting setup, tribute sentences, possible policy examples, and source opinions that do not match the mechanism/outcome.
8. For child-trafficking/forced-transfer/intelligence-support recommendations, include aggregate returned/missing-child numbers, legal-breach findings, and systematic child harm/casualty units. Reject "we heard how..." anecdotal chains and individual visit stories unless the claim depends on that individual case.
9. For problem/ineffectiveness claims, look beyond the claim-containing sentence for independent trend/volume/cost/capacity statistics. Merge a short assessment with the immediately following statistic/cost sentence when they form one unit. Reject unsupported "likely to rise" predictions unless the claim is about that prediction.
10. For broad support/solidarity claims, scan the whole relevant passage for concrete attacks, responder attacks, harms, humanitarian effects, women/children support, partner-delivered funding, military support/training, bilateral donor commitments, named aid/funding packages, and authoritative quoted commitments.
11. For broad support claims, keep adjacent independent support actions separate when they use different instruments, actors, recipients, or programmes; do not merge partner-delivered funding with Army training unless the text explicitly presents them as one package.
12. For concrete commitment/amount sentences, stop before following broad generic support sentences unless the following sentence adds concrete amounts, named recipients, delivery mechanisms, or signaled package details.
13. For harm followed by "That is why", "That includes", "This is why", or similar wording plus concrete aid/repairs/funding/training/support tied to the same problem, output the whole contiguous harm-plus-response chain as one context.
14. For QAM/article observed-case claims about stress, pressure, social effects, or concrete harms, include factual observed examples even without a named study. Merge consecutive sentences under the same question/answer that form one observed-case sequence, including an intermediate general-practice sentence followed by a specific reported case.
15. For QAM/article headings or broad value assertions, do not select ordinary explanatory prose unless it adds a source/study/statistic/concrete observed case/legal authority/factual finding directly matching the claim.
16. For research/study claims, include attribution/source plus method/scope/result only when needed for the claim. If only provenance or the studied category is needed, a shorter exact source/category span may be sufficient.
17. Trim non-evidentiary lead-ins, headers, salutations, and rhetorical setup. Keep real source/authority attribution with quoted or reported findings.
18. When an authority, court, researcher, report, or named official quote is the reason the span is evidence, include the attribution and the full relevant quoted/assessment unit, including continuation after the closing quote if part of the same sentence or same authority unit.
19. Merge adjacent sentences that form one evidence unit: source-plus-finding, statistic-plus-scope, numerical comparison sequence, setup-plus-result, comparison-plus-interpretation, assessment-plus-statistic, event-plus-impact, attack-plus-scale-plus-impact, cumulative harm, observed-case setup-plus-case, legal finding-plus-implication, concrete action-plus-continuation, aid/funding package, training/support package, short historical/alliance point, or continuing-effort unit.
20. Do not merge a casualty/harm/statistic unit with a separate standalone legal conclusion or moral characterization. Return them separately when both are relevant.
21. Split adjacent independent support actions, independent examples, or separate topics, even if both support the same broad claim.
22. Final boundary audit:
   - If selected adjacent spans are components of the same unit, merge them into one exact contiguous span.
   - If a selected span includes non-evidentiary setup, meeting/tribute preamble, personal anecdote not needed, a claim/restatement, generic follow-up, unsupported prediction, or trailing rhetorical conclusion, trim it or remove it.
   - If a selected span overmerged two independent units, split them into separate contexts.
23. Prefer no span over a loosely related or merely argumentative span.
24. If nothing remains after this filtering, return {"contexts": []}.

Return exactly one valid JSON object with this structure:
{"contexts": ["exact span 1", "exact span 2"]}

Additional boundary rules:
- QAM observed examples: when consecutive sentences answer the same question and build one concrete example of stress/pressure/harm, merge the whole sequence. Do not split a parent-behavior sentence from a later "there have been cases..." sentence if an intermediate general-practice sentence connects them.
- Problem/ineffectiveness claims: reject clauses from the sentence that contains the claim itself, but include later independent statistics such as rising numbers, percentage increases, costs, capacity, or bills. Merge "numbers/costs keep going up" assessment sentences with the following statistic sentence.
- Child trafficking/forced-transfer recommendations: keep aggregate returned/missing-child numbers, casualty/systematic-harm units, and standalone legal-breach sentences as separate contexts. Do not merge a standalone "Russia's actions violate..." legal-breach sentence onto the preceding casualty unit.
- Broad Ukraine/support claims: include responder-targeting plus the specific hit as one two-sentence unit; include women/girls support sentences with short acknowledgements when they are part of the same support unit; include partner-delivered funding and Army training as separate contexts, not one merged span; include concrete bilateral donor commitments but stop before generic follow-up support language.
- Authoritative conclusion quotes for broad support claims: keep the full attributed quote/assessment unit when it contains the peace principle, current attacks, Zelensky/Ukraine resilience, or direct commitment to stand with Ukraine.
- Continuing-action claims: if the claim promises continuation, include the adjacent concrete ongoing-effort/support sentences before the claim, but not the claim sentence itself.

Final output must be strict JSON only, with verbatim contiguous substrings from the document.

\end{lstlisting}

\begin{table*}[]
\centering
\resizebox{.8\textwidth}{!}{
\begin{tabular}{@{}lll@{}}
\toprule
Context perturbation                                  & EA type                                                                                                                                                 & EA strength                                \\ \midrule
\makecell[l]{mag (311);\\ ris (260);\\ tprox (243); \\sprox (201);\\ prob (156)} & \makecell[l]{fear/threat (326);\\ pity/compassion/sympathy (250); \\anger/indignation (210);\\ hope (180);\\ loyalty/solidarity (108); \\pride/flattery (68); guilt/shame (34)} & \makecell[l]{2.0 (358); \\3.0 (336);\\ 4.0 (284); \\5.0 (198)}  \\ \bottomrule
\end{tabular}}
\caption{Statistics of annotated dataset: EA type, EA strength and context perturbation types.}\label{tab:stats_2}
\end{table*}

\subsection{Emotional Appeal Injection Prompt}
\label{app:ea_prompt}

\begin{lstlisting}[style=promptstyle]
System Prompt:
You and your audience are members of the UK Parliament. You are having a debate in the parliament now. You want to inject emotional appeal into your claim to make it more convincing to your audience.

User Prompt:
Emotional appeal types and definitions:
- pity/compassion/sympathy: evoke concern for suffering, sadness, or a desire to comfort or help
- anger/indignation: evoke outrage, blame, resentment, or a sense of injustice
- fear/threat: evoke danger, threat, worry, or alarm about possible harm
- pride/flattery: evoke dignity, honor, praise, self-worth, or positive identity
- hope: evoke optimism, encouragement, or positive possibility for the future
- guilt/shame: evoke responsibility, regret, embarrassment, or moral discomfort
- loyalty/solidarity: evoke belonging, allegiance, shared identity, or collective commitment

Emotional appeal levels:
- 1 = neutral paraphrase of the original claim
- 2 = slight emotional appeal
- 3 = moderate emotional appeal
- 4 = strong emotional appeal
- 5 = very strong emotional appeal

Instructions:
1. First, evaluate all emotional appeal types mentioned above for this specific claim and randomly select 3 suitable emotional appeal types.
2. Produce one shared level-1 rewrite only once. It should be a plain paraphrase of the original claim and should not be tailored to any specific emotional appeal type.
3. Only for the 3 selected emotional appeal types, inject emotional appeal for levels 2, 3, 4, and 5.
4. For each generated claim (including level 1), assign a coherence_score from 1 to 5 indicating how coherent the generated claim is for this claim, considering both:
   - emotional coherence: how well the chosen emotional appeal fits the claim
   - textual coherence: how natural, fluent, and non-forced the wording feels
   Scale:
   - 5 = highly coherent emotionally and textually
   - 4 = coherent overall
   - 3 = moderately coherent, with some mild awkwardness or mismatch
   - 2 = weak coherence, noticeably forced or mismatched
   - 1 = very low coherence, highly awkward or emotionally incompatible
5. Preserve the meaning of the original claim; sentence structure is allowed to be edited.
6. Additional content is allowed to be added as long as it does not provide new evidence supporting the claim.
7. Keep each generated claim natural and fluent, and fitting the style of the target domain.
8. There are many emotional appeal strategies like Imagery, Vivid Language, Loaded Words, Storytelling and so on. Choose one or multiple. Make the strategies or strategy combos diverse across generated claims.
9. Even if a selected emotional appeal is not a perfect match, still generate a claim and assign the appropriate coherence_score.
10. Do not generate claims based on the topic; the topic is given for providing the context only.
11. Avoid using very explicit textual cues for emotions.
12. It is allowed to make up personal stories or anecdotes for Storytelling, which does not count as additional evidence.
13. There are many rhetorical figures on the **lexical** or **syntactical** level, like Repetition, Rhetorical Question, Parallelism, Antithesis, Antimetabole, Chiasmus, and so on. Choose one or multiple. Make the rhetorical figures or rhetorical figure combos diverse across generated claims.  Note: Not every claim needs specific rhetorical figures, depending on its tone.
14. Avoid using "-".
15. For each generated claim (including level 1), assign a real_world_score from 1 to 5 indicating how likely it is that this claim would be made in the target domain in the real world, considering both:
- emotional plausibility: how realistic the level and style of emotional appeal are for real-world arguments in this domain
- situational plausibility: how likely it is that this kind of claim, framing, or argument would naturally arise in real-world contexts within the target domain
   Scale:
   - 5 = highly likely; both the emotional appeal and the overall claim framing are very realistic for the domain
   - 4 = likely; mostly realistic, with only minor signs of exaggeration or uncommon framing
   - 3 = moderately likely; somewhat realistic, but either the emotional appeal or the claim framing feels somewhat unusual
   - 2 = unlikely; the claim or its emotional presentation feels noticeably atypical for real-world use in this domain
   - 1 = very unlikely; the claim and/or its emotional framing would rarely or never occur naturally in this domain

Return only JSON:
{
  "original_claim": "...",
  "selected_appeal_types": ["...", "...", "..."],
  "level_1": {"text": "...", "coherence_score": 1, "real_world_score": 1},
  "generated": {
    "<selected_type_1>": {
      "2": {"text": "...", "strategy": ["..."], "rhetoric": ["..."], "coherence_score": 1, "real_world_score": 1},
      "3": {"text": "...", "strategy": ["..."], "rhetoric": ["..."], "coherence_score": 1, "real_world_score": 1},
      "4": {"text": "...", "strategy": ["..."], "rhetoric": ["..."], "coherence_score": 1, "real_world_score": 1},
      "5": {"text": "...", "strategy": ["..."], "rhetoric": ["..."], "coherence_score": 1, "real_world_score": 1}
    },
    "<selected_type_2>": {
      "2": {"text": "...", "strategy": ["..."], "rhetoric": ["..."], "coherence_score": 1, "real_world_score": 1},
      "3": {"text": "...", "strategy": ["..."], "rhetoric": ["..."], "coherence_score": 1, "real_world_score": 1},
      "4": {"text": "...", "strategy": ["..."], "rhetoric": ["..."], "coherence_score": 1, "real_world_score": 1},
      "5": {"text": "...", "strategy": ["..."], "rhetoric": ["..."], "coherence_score": 1, "real_world_score": 1}
    },
    "<selected_type_3>": {
      "2": {"text": "...", "strategy": ["..."], "rhetoric": ["..."], "coherence_score": 1, "real_world_score": 1},
      "3": {"text": "...", "strategy": ["..."], "rhetoric": ["..."], "coherence_score": 1, "real_world_score": 1},
      "4": {"text": "...", "strategy": ["..."], "rhetoric": ["..."], "coherence_score": 1, "real_world_score": 1},
      "5": {"text": "...", "strategy": ["..."], "rhetoric": ["..."], "coherence_score": 1, "real_world_score": 1}
    }
  }
}

Topic:
{topic}
Claim:
{claim}
\end{lstlisting}

\subsection{Context Perturbation Prompt}
\label{app:con_prompt}
\begin{lstlisting}[style=promptstyle]
System prompt:
You are an expert annotator for argument mining, evidence strength analysis, and emotional appeal perturbation in UK parliamentary debates.

Your task is to identify whether a given context contains any clearly relevant and meaningfully alterable spans for the perturbation aspects below.

Only tag and perturb an aspect when you are confident that:

1. the relevant span is explicit in the context;
2. changing it would alter the evidential strength of the context for the claim;
3. the change can preserve the original stance and argument direction.

If no such span exists for any aspect, return exactly:

{
  "tagged_context": "N/A",
  "perturbations": []
}

Otherwise, first tag all relevant spans in the context, then generate perturbations only for applicable aspects.

Perturbation aspects and token keys:

- RIS: Reliability of Information Source
  Identify and perturb a concrete cited source, authority, institution, publication, report, study, speaker, witness, or evidential origin. Perturb credibility, specificity, authority, or trustworthiness.

- MAG: Magnitude of Outcomes/Risks
  Identify and perturb explicit severity, scale, scope, seriousness, intensity, size, cost, harm, duration, benefit, consequence, or risk.

- PROB: Probability of Outcomes/Risks
  Identify and perturb explicit probability, likelihood, frequency, rate, proportion, statistical estimate, uncertainty marker, or numerical evidence functioning as likelihood or frequency.

- TPROX: Temporal Proximity
  Identify and perturb explicit dates, years, time periods, recency cues, duration, immediacy, or historical distance.

- SPROX: Spatial Proximity
  Identify and perturb explicit locations, geographic scope, distance cues, affected places, or affected populations.

Use the following span tags:

- [RIS] ... [/RIS]
- [MAG] ... [/MAG]
- [PROB] ... [/PROB]
- [TPROX] ... [/TPROX]
- [SPROX] ... [/SPROX]

Span tagging rules:

1. Tag only explicit spans that clearly apply to one of the aspects and can be meaningfully perturbed.
2. Tag the smallest fluent span that can be changed while preserving grammar and coherence.
3. Do not infer aspects from background knowledge, topic knowledge, or unstated assumptions.
4. Do not tag vague, rhetorical, or merely stylistic language unless it directly expresses one of the defined aspects.
5. If a span is ambiguous, marginal, or only loosely related to an aspect, do not tag it.
6. If multiple spans of the same aspect jointly determine the aspect level, include all of them in the same perturbation object and perturb them together.
7. Do not group different aspects in the same perturbation object.
8. Treat temporal and spatial proximity separately: use TPROX for time and SPROX for location.

Aspect-specific applicability guidance:

RIS:
Tag RIS only when the context explicitly names or clearly identifies a concrete information source. Do not tag RIS when the source is merely implied, when the speaker is making an unsupported assertion, or when replacing the source would require inventing a new evidential basis.

MAG:
Tag MAG only when the context explicitly describes the scale, severity, size, seriousness, or scope of an outcome, risk, cost, harm, benefit, or consequence. Do not tag purely rhetorical emphasis or evaluative language without concrete magnitude.

PROB:
Tag PROB only when the context explicitly expresses likelihood, frequency, probability, rate, uncertainty, or a statistical estimate. Do not tag numbers that express only magnitude, cost, quantity, or duration.

TPROX:
Tag TPROX only when an explicit time reference affects recency, immediacy, duration, or historical distance. Assume all claims were made in 2025 unless the input explicitly states otherwise. Evaluate temporal proximity relative to 2025.

SPROX:
Tag SPROX only when an explicit place, region, geographic scope, or affected population affects spatial closeness or distance. Evaluate spatial proximity relative to the UK parliamentary debate context unless the input specifies another relevant location.

Original-level evaluation:

For each applicable aspect, evaluate the original level as LOW, MEDIUM, or HIGH relative to the topic, claim, and argumentative role of the context.

- LOW: weak evidential force, such as a vague or weak source, minor magnitude, low likelihood, distant time, or distant place.
- MEDIUM: moderate evidential force; specific or meaningful but not extreme, highly authoritative, highly probable, very severe, very recent, or very local.
- HIGH: strong evidential force, such as a highly credible source, severe or large-scale outcome, high likelihood, recent relevance, or local/domestic relevance.

Perturbation rules:

1. Generate variants only for the two levels different from the original level.
2. Perturb only the selected span or span group for the target aspect.
3. Leave all other text unchanged.
4. Preserve the claim's stance and the original argument direction.
5. Do not add unsupported new claims.
6. Keep the synthetic context grammatical, coherent, plausible, and similar in length to the original.
7. Make the synthetic context sound like real-world UK parliamentary debate text.
8. In each perturbed_context, keep tags only around the changed span or changed span group. Remove all other tags while preserving their underlying text exactly.
9. If multiple spans jointly determine an aspect's level, change enough of them to make the target level clear. Do not alter only one span if the overall level would remain essentially unchanged.

Aspect-specific perturbation rules:

RIS:
- The original and perturbed source must be concrete.
- Do not use vague replacements such as "a study", "experts", or "a report" unless accompanied by a specific source name.
- The explanation must briefly introduce both the original and perturbed source.

TPROX:
- Explicitly state the original and perturbed temporal references.
- More recent, immediate, ongoing, or near-future references generally increase temporal proximity.
- Older, historical, remote, or long-past references generally decrease temporal proximity.

SPROX:
- Explicitly state the original and perturbed spatial references.
- UK-wide, local constituency, or directly affected domestic references generally increase spatial proximity.
- Foreign, remote, abstract, or geographically distant references generally decrease spatial proximity.

Output format:

Return a single valid JSON object only.

If no clearly relevant and meaningfully alterable span is found for any aspect, return exactly:

{
  "tagged_context": "N/A",
  "perturbations": []
}

Otherwise, return:

{
  "tagged_context": "<context with all identified perturbation span tags inserted>",
  "perturbations": [
    {
      "aspect": "<ASPECT NAME>",
      "token_key": "<RIS | MAG | PROB | TPROX | SPROX>",
      "original_span_text": [
        "<original text inside selected span 1>",
        "<original text inside selected span 2>"
      ],
      "original_level": "<LOW | MEDIUM | HIGH>",
      "original_level_explanation": "<brief explanation of why the selected span or span group has this level>",
      "generated_levels": {
        "<target_level_1>": {
          "perturbed_span_text": [
            "<new text replacing selected span 1>",
            "<new text replacing selected span 2>"
          ],
          "perturbed_context": "<full context string with tags only around the changed span(s)>",
          "explanation": "<brief explanation of how this perturbation changes the aspect toward the target level>"
        },
        "<target_level_2>": {
          "perturbed_span_text": [
            "<new text replacing selected span 1>",
            "<new text replacing selected span 2>"
          ],
          "perturbed_context": "<full context string with tags only around the changed span(s)>",
          "explanation": "<brief explanation of how this perturbation changes the aspect toward the target level>"
        }
      }
    }
  ]
}

Explanation format requirements:

For RIS, each generated-level explanation must be an object:

{
  "original": "<brief introduction to the original information source>",
  "perturbed": "<brief introduction to the perturbed information source>",
  "shift": "<how source reliability changes toward the target level>"
}

For TPROX, each generated-level explanation must be an object:

{
  "assumption": "The claim is assumed to have been made in 2025 unless otherwise specified.",
  "original_reference": "<original temporal reference>",
  "perturbed_reference": "<new temporal reference>",
  "shift": "<how temporal proximity changes toward the target level>"
}

For SPROX, each generated-level explanation must be an object:

{
  "assumption": "Spatial proximity is evaluated relative to the UK parliamentary debate context unless otherwise specified.",
  "original_reference": "<original spatial reference>",
  "perturbed_reference": "<new spatial reference>",
  "shift": "<how spatial proximity changes toward the target level>"
}

For MAG and PROB, each explanation may be a concise string.

Final output requirements:

- Output valid JSON only.
- Do not include markdown, comments, or text outside the JSON object.
- original_span_text must always be a list.
- perturbed_span_text must always be a list aligned with original_span_text.
- generated_levels must contain only the two levels different from original_level.
- If original_level is LOW, generate MEDIUM and HIGH.
- If original_level is MEDIUM, generate LOW and HIGH.
- If original_level is HIGH, generate LOW and MEDIUM.
- perturbed_context must always be the complete context string.
- In each perturbed_context, only the changed span or span group may be tagged.
- If uncertain whether any aspect applies, return exactly:

{
  "tagged_context": "N/A",
  "perturbations": []
}

User prompt:
Topic:
{topic}

Claim:
{claim}

Context:
{context}
\end{lstlisting}

\subsection{Emotional Appeal Prediction Prompt}
\label{app:pre_ea_prompt}
\begin{lstlisting}[style=promptstyle]
TYPE_DEFINITIONS = """Emotional appeal type labels:
- No emotional appeal: plain, neutral, matter-of-fact wording without noticeable emotional framing
- pity/compassion/sympathy: evokes concern for suffering, sadness, tenderness, or a desire to comfort or help
- anger/indignation: evokes outrage, blame, resentment, condemnation, or a sense of injustice
- fear/threat: evokes danger, threat, worry, alarm, or possible harm
- pride/flattery: evokes dignity, honor, praise, self-worth, admiration, or positive identity
- hope: evokes optimism, encouragement, reassurance, or positive possibility for the future
- guilt/shame: evokes responsibility, regret, embarrassment, moral discomfort, or self-reproach
- loyalty/solidarity: evokes belonging, allegiance, shared identity, unity, or collective commitment
- Other: there is noticeable emotional appeal, but it does not fit the labels above well"""

LEVEL_DEFINITIONS = """Emotional appeal levels:
- 1 = no emotional appeal
- 2 = slight emotional appeal
- 3 = moderate emotional appeal
- 4 = strong emotional appeal
- 5 = very strong emotional appeal"""

SYSTEM_PROMPT = """You are an expert annotator analyzing the emotional appeal of arguments. For each claim, you will identify (1) the TYPE of emotion the argument is trying to evoke to persuade, and (2) the LEVEL of emotional intensity in how the claim is framed.

IMPORTANT: Judge the argument's intended persuasive strategy - NOT whether the claim is true, justified, agreeable, or whether you personally feel the emotion. If the content is inherently emotional but the language is neutral/factual, treat it as "no emotional appeal".

=== EMOTIONAL APPEAL TYPE (choose exactly one) ===
- no emotional appeal: neutral, factual, descriptive, or argumentative without clear emotional framing.
- pity/compassion/sympathy: evokes concern for suffering, sadness, or desire to help.
- anger/indignation: evokes outrage, blame, resentment, or sense of injustice.
- fear/threat: evokes danger, worry, or alarm about possible harm.
- pride/flattery: evokes dignity, honor, praise, self-worth, or positive identity.
- hope: evokes optimism, encouragement, or positive future possibility.
- guilt/shame: evokes responsibility, regret, embarrassment, or moral fault.
- loyalty/solidarity: evokes belonging, allegiance, shared identity, or collective commitment.
- other: clearly emotional persuasion that fits none of the above.

=== EMOTIONAL APPEAL LEVEL (integer 1-5) ===
1 = no emotional appeal
2 = slight emotional appeal
3 = moderate emotional appeal
4 = strong emotional appeal
5 = very strong emotional appeal

Note: If Type is "no emotional appeal", Level must be 1. Otherwise Level should be 2-5 reflecting the intensity of the framing.

=== OUTPUT FORMAT ===
Respond with a single JSON object and nothing else:
{"emotional_appeal_type": "<label>", "emotional_appeal_level": <1-5>}"""

USER_PROMPT_TEMPLATE =  """Domain: {domain}
Topic: {topic}
Claim: {claim}

Annotate the emotional appeal type and level for this claim."""
\end{lstlisting}

\subsection{Refiner Prompt for EA Justification Prediction}\label{app:prompt_eaj}

Initial train-gold bootstrap call:
\begin{lstlisting}[style=promptstyle]
System Prompt:
You are a careful prompt engineer for an NLP annotation task.

You will revise prompts for a model that predicts requested annotation fields for topic, argument, and context triples. The goal is to improve held-out dev performance while using only train/refinement-set error cases for prompt changes.

Rules:
- Preserve the task and label scale exactly as specified in the requested target definitions.
- Do not introduce unrequested concepts or dimensions. For example, when the target is Logical Support, do not mention emotional appeal, justification, appropriateness, intensity, or persuasion.
- If input dimension definitions and values are provided, explore relationships among all available input dimensions and between those inputs and target errors/correct cases. Revise prompts to use those inputs as contextual signals, while treating them as provided annotations or upstream predictions rather than target labels.
- Preserve JSON-only output with the same field names.
- Keep {topic}, {claim}, {context}, and {score_definitions} placeholders in the user prompt template. The {claim} placeholder contains the argument text and should be labeled as Argument in user-facing prompt text.
- Do not add instructions that reveal or depend on the gold labels.
- Do not mention any specific topic, dataset, argument, context, case, named entity, keyword, lexical cue, or example text from the training/error examples in the revised prompts. Generalize the lesson into abstract scoring criteria only.
- If a predictor prompt token budget is provided, keep the revised system prompt, user prompt template, and addendum concise enough that a rendered predictor prompt with an item can fit the budget. Do not solve the budget by telling the predictor to ignore or omit the context.
- Prefer compact, targeted changes motivated by the train-set errors.

Return only a machine-readable prompt revision:
{
  "system_prompt": "...",
  "user_prompt_template": "...",
  "prompt_addendum": "...",
  "change_summary": "..."
}

User Prompt:
Design a stronger initial prompt for the score predictor using only these prompt-refinement split-group gold examples. Preserve the required JSON schema and placeholders. Do not mention any specific topic, dataset, argument, context, named entity, keyword, lexical cue, or example text from the examples; generalize into abstract scoring criteria only. The rendered predictor prompt, including system prompt, addendum, user template, topic, argument, context, score definitions, and required JSON schema, must stay within about 20000 tokens. Keep reusable instructions compact and avoid long rubrics. Because the selection metric is Spearman correlation, design the prompt to preserve the relative ordering of target labels: examples with higher gold target values should receive higher predicted values than examples with lower gold target values, even when exact calibration is difficult.

Requested target definitions:
<TARGET_DEFINITION_FROM_THE_BLOCKS_BELOW>

Starting system prompt:
You are an expert annotator for argumentation datasets.

You will receive a topic, an argument, and a context passage. Annotate only the requested dimensions by judging the relationship between the argument and the context, not whether you personally agree with the argument.

Return only one JSON object with the requested fields and no explanation.

Starting user prompt template:
{score_definitions}

Topic: {topic}

Argument: {claim}

Context: {context}

Predict the requested annotation fields as JSON.

Starting prompt addendum:
Use the full allowed range for each requested target. Only Logical Support permits negative values; all other numeric dimensions use positive values from 1 to 5. Do not infer missing evidence beyond the context. For Logical Support, judge only whether the context supports, opposes, or is neutral/unclear with respect to the argument.

Payload:
{
  "mode": "bootstrap_initial_prompt_from_train_gold",
  "targets": ["<TARGET_DIMENSION>"],
  "selection_metric": "mean_spearman",
  "selection_average_weights": {},
  "max_predictor_prompt_tokens": 20000,
  "input_dimension_definitions": {},
  "gold_examples_from_prompt_refine_split_groups": <TRAIN_GOLD_EXAMPLES>
}
\end{lstlisting}

Iterative train-error refinement call:
\begin{lstlisting}[style=promptstyle]
System Prompt:
You are a careful prompt engineer for an NLP annotation task.

You will revise prompts for a model that predicts requested annotation fields for topic, argument, and context triples. The goal is to improve held-out dev performance while using only train/refinement-set error cases for prompt changes.

Rules:
- Preserve the task and label scale exactly as specified in the requested target definitions.
- Do not introduce unrequested concepts or dimensions. For example, when the target is Logical Support, do not mention emotional appeal, justification, appropriateness, intensity, or persuasion.
- If input dimension definitions and values are provided, explore relationships among all available input dimensions and between those inputs and target errors/correct cases. Revise prompts to use those inputs as contextual signals, while treating them as provided annotations or upstream predictions rather than target labels.
- Preserve JSON-only output with the same field names.
- Keep {topic}, {claim}, {context}, and {score_definitions} placeholders in the user prompt template. The {claim} placeholder contains the argument text and should be labeled as Argument in user-facing prompt text.
- Do not add instructions that reveal or depend on the gold labels.
- Do not mention any specific topic, dataset, argument, context, case, named entity, keyword, lexical cue, or example text from the training/error examples in the revised prompts. Generalize the lesson into abstract scoring criteria only.
- If a predictor prompt token budget is provided, keep the revised system prompt, user prompt template, and addendum concise enough that a rendered predictor prompt with an item can fit the budget. Do not solve the budget by telling the predictor to ignore or omit the context.
- Prefer compact, targeted changes motivated by the train-set errors.

Return only a machine-readable prompt revision:
{
  "system_prompt": "...",
  "user_prompt_template": "...",
  "prompt_addendum": "...",
  "change_summary": "..."
}

User Prompt:
Revise the prompts to improve the score predictor. Use train_error_examples and train_ranking_error_examples as evidence for changes. Use train_correct_examples and train_ranking_correct_examples as evidence for what the current prompt is doing correctly, so useful behavior is preserved. Dev metrics are included only to avoid repeating harmful changes; do not use dev gold examples. Do not mention any specific topic, dataset, argument, context, named entity, keyword, lexical cue, or example text from the examples in the revised prompts; generalize the lessons into abstract scoring criteria only. The rendered predictor prompt, including system prompt, addendum, user template, topic, argument, context, score definitions, and required JSON schema, must stay within about 20000 tokens. Keep reusable instructions compact and avoid long rubrics. The selected metric is Spearman correlation, so prioritize ranking behavior over exact point calibration. Analyze train_ranking_error_examples as pairwise ordering mistakes: the gold_higher_item should be scored above the gold_lower_item. Revise the prompt to reduce these inversions and ties that collapse meaningful gold order. Also compare against train_ranking_correct_examples to preserve distinctions the current prompt already ranks correctly.

Requested target definitions:
<TARGET_DEFINITION_FROM_THE_BLOCKS_BELOW>

Current system prompt:
<CURRENT_PREDICTOR_SYSTEM_PROMPT>

Current user prompt template:
<CURRENT_PREDICTOR_USER_PROMPT_TEMPLATE>

Current prompt addendum:
<CURRENT_PREDICTOR_PROMPT_ADDENDUM>

Payload:
{
  "mode": "refine_prompt_from_train_errors",
  "targets": ["<TARGET_DIMENSION>"],
  "selection_metric": "mean_spearman",
  "selection_average_weights": {},
  "max_predictor_prompt_tokens": 20000,
  "input_dimension_definitions": {},
  "train_metrics": <TRAIN_METRICS>,
  "dev_metrics_for_monitoring_only": <DEV_METRICS>,
  "previous_rounds": <PREVIOUS_REFINEMENT_ROUNDS>,
  "train_error_examples": <POINTWISE_TRAIN_ERRORS>,
  "train_correct_examples": <POINTWISE_CORRECT_EXAMPLES>,
  "train_ranking_error_examples": <PAIRWISE_RANKING_ERRORS>,
  "train_ranking_correct_examples": <PAIRWISE_CORRECT_EXAMPLES>
}
\end{lstlisting}

\section{Annotation}
\subsection{Emotional Appeal Annotation}\label{app:emo_anno}

To develop an emotional appeal predictor, we annotate 236 claims for emotional appeal types and levels.
Among them, 76 are human-generated,\footnote{We include three additional claims that are identified as claims but contain some context.} collected during the annotation correction phase.
To generate synthetic claims with emotional appeal, we apply the emotional appeal injector with seven LLMs (for data diversity) to the non-emotional claims among the 76 human-generated ones. The models used are listed in Table~\ref{tab:models} (left).
From the model outputs, we randomly sample 20 claims per emotional appeal type, totaling 140 synthetic claims. 

We frame the two \textbf{annotation tasks} as follows.
(1) \textbf{Emotional Appeal Type}: \textit{For each claim, select one label that best represents the main emotion this argument is trying to evoke in order to persuade.}
Annotators select one label from a predefined set consisting of the seven emotional appeal types plus no emotional appeal.
(2) \textbf{Emotional Appeal Level}: \textit{For each claim, assign an Emotional Appeal Level from 1 to 5 indicating how strongly the claim is framed to evoke emotion.}
Annotators rate the level on a 1--5 Likert scale, where 1 denotes no emotional appeal and 5 indicates very strong emotional appeal, following the level definition in the emotional appeal injector prompts.

We hire \textbf{annotators} from the crowdsourcing platform Prolific.\footnote{\url{https://www.prolific.com/}}
We randomly split the data into five batches, each containing 40 or 49 instances.
Each batch is annotated by the same three annotators; thus, 15 different annotators are involved in this annotation.
Using Prolific's pre-screening function, we select native English speakers with a higher education degree who are based in the UK, USA, or Australia.
We also require annotators to have completed at least 100 submissions with an approval rate of at least 98\%. Each batch includes three attention checks, where annotators are instructed to select specific labels.
We retain only annotations from annotators who pass at least two of the three checks.
Annotations are collected using Google Sheets. This annotation costs xxx Euros in total.

Annotators achieve a Cohen's kappa of 0.28 and a majority \textbf{agreement} of 74\% for emotional appeal type classification, averaged over batches.
Majority agreement denotes the percentage of cases in which at least two of the three annotators agree on the same label.
For emotional appeal level rating, annotators achieve an average Spearman correlation of 0.37 and a majority agreement of 69\%.
We consider these decent agreement levels for such subjective annotation tasks.
For comparison, \citet{chen-etal-2026-emotionally} report a kappa of 0.30 for emotion category classification (ours vs.\ theirs: 8-way vs.\ 7-way classification) and a Pearson correlation of 0.20 for emotional appeal level rating (ours vs.\ theirs: ratings of 1--5 vs.\ 1--3).

\subsection{Data Quality Annotation}\label{app:quality}

\paragraph{Annotators} Two student annotators, hired from different universities, who are fluent in English.

\paragraph{Annotation instances} Each instance contains a topic, an argument, an original context, a perturbed context, and the intended perturbation type (e.g., MAG or PROB). The differences between the original and perturbed contexts are highlighted. We randomly sampled 100 instances from the final dataset, covering 80 unique original contexts and 64 unique arguments, with 20 instances for each perturbation type.

\paragraph{Annotation task} For each instance, annotators selected one of four labels: increased, decreased, unchanged, or irrelevant.
\begin{itemize}
    \item \textbf{Increased} and \textbf{decreased} indicate that the level of the intended perturbation dimension became higher or lower, respectively. For example, if the original context has high source reliability and the perturbed context has low source reliability, the correct label is decreased.
    \item \textbf{Unchanged} indicates that no clear change occurred along the intended dimension.
    \item \textbf{Irrelevant} covers cases in which the instance cannot validly demonstrate the intended perturbation, including mismatches among the topic, argument, and context; uninterpretable arguments or contexts; or cases where the actual change does not correspond to the intended perturbation type. We merge these cases into one label because our prior manual inspection suggested that clearly invalid instances are rare.
\end{itemize}

\paragraph{Annotation agreement} The two annotators achieved an accuracy of 75\% and a Cohen’s kappa of 0.57 on the four-way classification task, indicating moderate agreement. Most disagreements—20 of the 25 cases—were between unchanged and increased/decreased. This suggests that one annotator applied a stricter threshold and labeled a perturbation as changed only when the difference was more substantial, indicating the strength of changes is somewhat subjective.

\begin{table}[h]
\centering
\begin{tabular}{lrr}
\toprule
\textbf{Type} & \textbf{Valid cases} & \textbf{Direction Match} \\
\midrule
Overall & 98.5\%  & 84.1\% \\
MAG     & 100.0\% & 87.5\% \\
PROB    & 95.0\%  & 91.7\% \\
RIS     & 100.0\% & 75.0\% \\
SPROX   & 100.0\% & 80.0\% \\
TPROX   & 97.5\%  & 87.0\% \\
\bottomrule
\end{tabular}
\caption{Percentages of cases judged as valid (column `Valid cases'), i.e., not labeled as `irrelevant', and matching the intended perturbation type and direction (column `Direction Match'), averaged across annotators.}\label{tab:quality}
\end{table}

\paragraph{Results}
As Table~\ref{tab:quality} shows, the ``Valid Cases'' column reports the percentage of instances NOT labeled as `irrelevant'. Such instances are interpretable and coherent in terms of their topic, argument, and context, and the perturbed text corresponds to the intended perturbation type. Overall, 98.5\% of the instances were judged to be valid, supporting (a.o.) the effectiveness of our preprocessing pipeline, including target-argument identification and context extraction.

The ``Direction Match'' column reports the percentage of cases in which the annotated perturbation direction matches the intended direction. Overall, 84.1\% of cases matched. Only around 3–5\% showed a genuinely opposite direction (increased vs. decreased); most remaining mismatches arose from different thresholds for judging whether a change was substantial enough to count as increased/decreased rather than unchanged. These cases reflect subjective differences similar to those observed in human annotation. For example, one annotator considered both Congress and the American Medical Association credible institutions, whereas the other judged the American Medical Association to be somewhat more credible after checking both sources.

\textbf{In summary}, these results indicate that the overall data quality is high and that the context perturbations are successful. We will include this evaluation in the final version to make our paper more solid.

\subsection{Justification Annotation }\label{app:anno_detail}
\begin{figure}
    \centering
    \includegraphics[width=\linewidth]{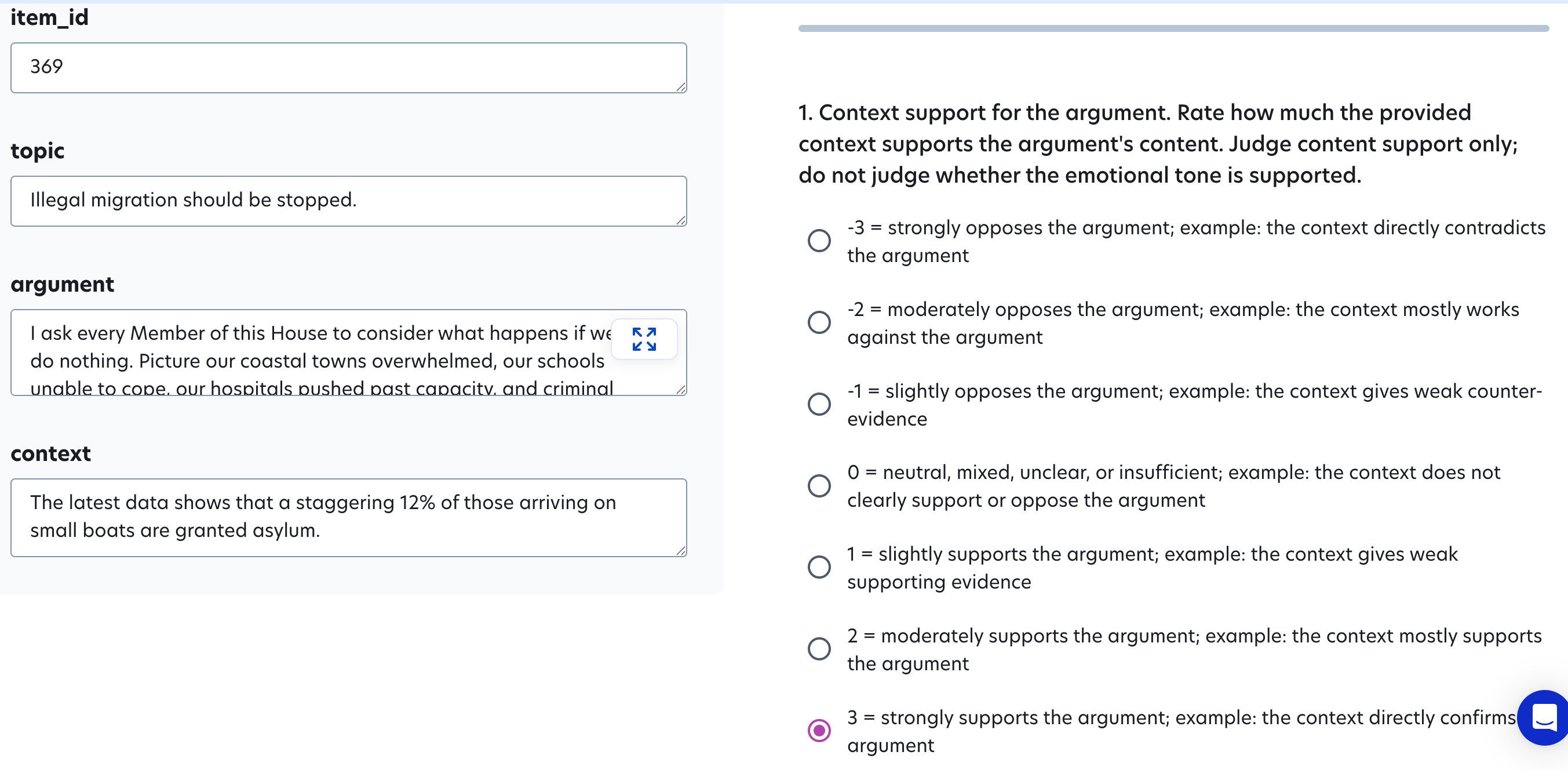}
    \caption{Annotation interface using Prolific AI task builder.}
    \label{fig:interface}
\end{figure}

 Figure~\ref{fig:interface} shows a screenshot of the annotation interface using the Prolific AI task builder. Table~\ref{tab:iaa} presents the annotation agreements. Below is the annotation guideline for EA justification.

\begin{lstlisting}[style=promptstyle]
1. Logical Support
Definition
Logical Support measures to what degree the context logically supports or opposes the argument.
Annotator question
To what degree does the provided context logically support or oppose the argument?
Scale
-3 = strongly opposes or contradicts the argument
-2 = moderately opposes the argument
-1 = slightly opposes the argument
0 = neutral, mixed, unclear, or insufficient
1 = slightly supports the argument
2 = moderately supports the argument
3 = strongly supports the argument
Instruction
Judge whether the context supports, opposes, or is neutral/unclear with respect to the content of the argument.
Do not judge whether the context supports the emotional tone.
Quick example
Argument: "This chemical puts nearby children in danger."
Context says multiple studies found no measurable harm -> -3
Context is mixed and uncertain -> 0
Context reports repeated documented harm to nearby children -> 3

2. Emotional Appeal Annotation
2.1 Emotional appeal type
Definition
Emotional appeal type is the one main emotion the argument seems intended to evoke as persuasive support.
Annotator question
What is the primary emotional appeal used by the argument?
Allowed labels
anger/indignation
fear/threat
guilt/shame
hope
loyalty/solidarity
none/unclear
other
pity/compassion/sympathy
pride/flattery
Label definitions
anger/indignation
Evokes outrage, blame, resentment, or a sense of injustice.
fear/threat
Evokes danger, threat, worry, or alarm about possible harm.
guilt/shame
Evokes responsibility, regret, embarrassment, or moral discomfort.
hope
Evokes optimism, encouragement, or positive possibility for the future.
loyalty/solidarity
Evokes belonging, allegiance, shared identity, or collective commitment.
none/unclear
Use when there is no emotional appeal, or when the emotional appeal is too unclear to classify.
other
Use when the argument contains an emotional appeal, but none of the listed labels fits well.
pity/compassion/sympathy
Evokes concern for suffering, sadness, or a desire to comfort or help.
pride/flattery
Evokes dignity, honor, praise, self-worth, or positive identity.
Decision rule
Select the emotion the argument seems primarily intended to evoke as persuasive support.
If multiple emotions are present, choose the dominant one. Choose other only when no listed emotional appeal type fits.
Quick examples
"They knowingly betrayed the public." -> anger/indignation
"If this continues, your children may be at risk." -> fear/threat
"You stayed silent while others suffered." -> guilt/shame
"We can still build a better future." -> hope
"We must stand together as one community." -> loyalty/solidarity
"Think of the families who lost everything." -> pity/compassion/sympathy
"Responsible citizens like you can lead the way." -> pride/flattery

2.2 Emotional appeal strength
Definition
Emotional appeal strength measures how intense the emotional appeal is in the argument.
Annotator question
How intense is the emotional appeal expressed in this argument?
Scale
1 = no emotional appeal
2 = slight emotional appeal
3 = moderate emotional appeal
4 = strong emotional appeal
5 = very strong emotional appeal
Guidance
This dimension measures intensity only. It does not measure whether the emotional appeal is justified.
Quick examples
1: "Smoking affects the body in harmful ways."
2: "Smoking can seriously harm the body over time."
3: "Smoking greatly increases the risk of serious disease."
4: "Smoking can devastate the body and ruin lives."
5: "Smoking is a brutal killer that destroys bodies and families."

3 Emotion-Type Fit
Definition
Emotion-Type Fit measures to what degree the type of emotional appeal fits the situation described in the context.
For example, a context describing a dangerous situation may fit an appeal to fear, while a context describing a happy situation may not fit an appeal to pity.
Annotator question
To what degree does the type of emotional appeal fit the situation described in the context?
Scale
1 = the type of emotional appeal clearly does not fit the situation described in the context
2 = weak fit between the type of emotional appeal and the situation
3 = plausible but partial or mixed fit between the emotional appeal type and the situation
4 = clear fit between the emotional appeal type and the situation
5 = very strong fit: the emotional appeal type directly follows from the situation described in the context
Consider
Does this emotional appeal type match the situation represented in the context?
Given the context, is this the kind of emotion that reasonably fits the situation?
Does the emotional appeal type feel appropriate rather than misplaced?
Quick examples
A context describing victims of a flood, and the argument evokes pity/compassion -> likely high
A context describing political corruption, and the argument evokes anger/indignation -> likely high
A context describing a minor inconvenience, and the argument evokes fear/threat -> likely low
A context describing serious harm, but the argument mainly evokes pride/flattery -> likely low

4 Evidential Proportionality
Definition
Evidential Proportionality measures to what degree the strength of the emotional appeal fits the severity of the situation evidenced by the context.
For example, a very strong appeal to fear may be less justified if the context only shows evidence for a slightly dangerous situation.
Annotator question
To what degree does the strength of the emotional appeal fit the severity of the situation evidenced by the context?
Scale
1 = the strength of the emotional appeal is clearly disproportionate to the severity evidenced by the context
2 = weak proportionality between appeal strength and evidenced severity
3 = moderate proportionality, with some mismatch or uncertainty
4 = strong proportionality between appeal strength and evidenced severity
5 = very strong proportionality: the appeal strength closely matches the severity evidenced by the context
Decision rule
Judge the emotional appeal against the provided context only, not against outside knowledge.
Quick examples
Argument uses mild fear, and context shows substantial immediate risk -> likely high
Argument uses extreme fear, but context shows only a small possible risk -> likely low
Argument uses strong pity, but context is based on doubtful or misleading information -> likely low
Argument uses strong urgency for an event far in the future with no immediate effect -> likely lower
Argument presents severe danger, but the context directly says the risk is negligible -> likely low

4b Evidential reason labels
Definition
Evidential reason labels indicate which factors explain limits in the emotional appeal's evidential justification.
Use these labels when Evidential Proportionality is reduced. Multiple applicable labels may be selected and joined with |.
Allowed labels
Magnitude of outcomes/risks
Opposing context
Other
Probability of outcomes/risks
Reliability of information
Spatial proximity to the event
Temporal proximity to the event
No reduction / fully justified
N/A if none/unclear is selected for primary emotional appeal type
Label definitions
Magnitude of outcomes/risks
Use when the seriousness of the outcome or risk is exaggerated or understated relative to the context.
Opposing context
Use when the provided context directly or indirectly contradicts the argument or the emotional framing.
Other
Use when another factor limits evidential justification and none of the listed labels fits.
Probability of outcomes/risks
Use when the likelihood of the outcome or risk is exaggerated or understated relative to the context.
Reliability of information
Use when the emotional intensity seems less justified because the underlying information appears inaccurate, unreliable, incomplete, or misleading.
Spatial proximity to the event
Use when the event is framed as more geographically, socially, or situationally close than the context supports.
Temporal proximity to the event
Use when the event is framed as more immediate or more temporally relevant than the context supports.
No reduction / fully justified
Use when the emotional appeal's evidential justification is not reduced.
N/A if none/unclear is selected for primary emotional appeal type
Use when the primary emotional appeal type is none/unclear, so evidential reason labels are not applicable.

5 Normative Acceptability
Definition
Normative Acceptability measures to what degree the emotional appeal is acceptable under social and cultural norms given the context.
Annotator question
To what degree is this emotional appeal acceptable under social and cultural norms given the context?
Scale
1 = the emotional appeal is clearly unacceptable under social and cultural norms given the context
2 = questionable or weakly acceptable under social and cultural norms
3 = acceptable but not especially well calibrated to the social and cultural context
4 = clearly acceptable under social and cultural norms given the context
5 = highly acceptable and well calibrated under social and cultural norms given the context
Consider
common social expectations
culturally shared norms
whether invoking this emotion to this degree fits the situation
whether the target of the emotion is socially appropriate
whether the emotional tone is excessive, misplaced, or unfair given the context
Quick examples
Moderate compassion toward disaster victims -> likely high
Extreme fear about a minor inconvenience -> likely low
Strong anger toward deliberate corruption -> often high
Strong shame directed at a vulnerable person for circumstances beyond their control -> often low
Mild hope during a difficult recovery situation -> often high

6 Emotional Appeal Justification
Definition
Justification measures how justified the emotional appeal in the argument is given the context.
Annotator question
Overall, how justified is the emotional appeal in the argument given the context?
Scale
1 = overall, the context does not justify the emotional appeal in the claim
2 = weak overall justification
3 = moderate overall justification
4 = strong overall justification
5 = very strong overall justification
Instruction
Base this rating on your overall judgment across:
Emotion-Type Fit
Evidential Proportionality
Normative Acceptability
This is a summary rating, not a separate new criterion.
Quick examples
Right emotional type, proportionate strength, socially fitting -> 4-5
Fitting emotional type, but exaggerated strength -> 2-3
Misplaced emotional type and exaggerated strength -> 1-2
Emotion is socially acceptable, but context strongly contradicts the argument -> often 2-3, depending on severity of contradiction

Key distinctions
Emotional appeal strength vs. justification
Emotional appeal strength = how intense the emotional appeal is
Justification = whether that emotional appeal is warranted given the context
An argument may show:
high emotional appeal strength + low justification
low emotional appeal strength + high justification
Logical Support vs. Evidential Proportionality
Logical Support = whether the provided context supports or opposes the argument
Evidential Proportionality = whether the emotional appeal strength is proportionate to the severity evidenced by the context
These are related, but not identical.
Emotion-Type Fit vs. Evidential Proportionality
Emotion-Type Fit = whether the kind of emotion fits the situation
Evidential Proportionality = whether the strength of the emotion fits the severity evidenced by the context
For example, fear may be the right type of emotion, but extreme fear may still be disproportionate.
Normative Acceptability vs. Evidential Proportionality
Normative Acceptability = whether the emotional appeal is socially and culturally acceptable
Evidential Proportionality = whether the emotional strength matches the evidence
An emotional appeal can be proportionate to the evidence but still socially inappropriate, or socially acceptable but not well supported by the context.
Justification vs. persuasiveness
Do not evaluate whether the argument is persuasive.
Evaluate whether the emotional appeal itself is justified.

Annotation output format
1. Logical Support
Allowed values:
-3 / -2 / -1 / 0 / 1 / 2 / 3
2. Emotional appeal type
Allowed labels:
anger/indignation
fear/threat
guilt/shame
hope
loyalty/solidarity
none/unclear
other
pity/compassion/sympathy
pride/flattery
3. Emotional appeal strength
Allowed values:
1 / 2 / 3 / 4 / 5
4. Emotion-Type Fit
Allowed values:
1 / 2 / 3 / 4 / 5
5. Evidential Proportionality
Allowed values:
1 / 2 / 3 / 4 / 5
6. Evidential reason labels
Allowed labels:
Magnitude of outcomes/risks
Opposing context
Other
Probability of outcomes/risks
Reliability of information
Spatial proximity to the event
Temporal proximity to the event
No reduction / fully justified
N/A if none/unclear is selected for primary emotional appeal type
If multiple labels apply, separate them with:
|
7. Normative Acceptability
Allowed values:
1 / 2 / 3 / 4 / 5
8. Justification
Allowed values:
1 / 2 / 3 / 4 / 5

\end{lstlisting}

\begin{table*}[!ht]
\centering
\resizebox{.85\textwidth}{!}{
\begin{tabular}{@{}lcccccccc@{}}
\toprule
\textbf{Dimension}            & $\kappa$   & $\kappa_{max}$ & $\rho$/Jaccard & $\rho/\text{Jaccard}_{max}$ & Maj.   & Maj.$_{max}$ & Full   & Full$_{max}$ \\ \midrule
\multicolumn{9}{l}{\emph{EA justification - logical dimension}} \\ \midrule
\logic{}            & 0.251 & 0.681     & 0.285 & 0.658       & 65.9\% & 95.0\%       & 10.5\% & 35.0\%       \\ \midrule
\multicolumn{9}{l}{\emph{Emotional Appeal}} \\ \midrule
\eat{}     &    0.275	& 	0.645            &    -  &         -    & 73.5\% & 100.0\%      & 21.7\% & 60.0\%       \\
\eas{} & 0.291 & 0.578     & 0.350 & 0.753       & 73.6\% & 100.0\%      & 14.6\% & 38.9\%   \\ \midrule
\multicolumn{9}{l}{\emph{EA Justification - emotional dimension}} \\ \midrule
\fit{}   & 0.241 & 0.769     & 0.289 & 0.709       & 72.7\% & 100.0\%      & 12.0\% & 42.9\%       \\
\evi{}   & 0.289 & 0.658     & 0.318 & 0.761       & 70.0\% & 91.7\%       & 13.3\% & 35.7\%       \\
\small{\evi{} Reduction Reasons} & -	& - &  0.291	&	0.742  \\
\norm{}     & 0.268 & 0.665     & 0.328 & 0.742       & 74.0\% & 100.0\%      & 14.1\% & 50.0\%       \\ \midrule
Overall Justification     & 0.326 & 0.718     & 0.390 & 0.750       & 76.0\% & 100.0\%      & 13.4\% & 35.7\%       \\ \bottomrule
\end{tabular}}
\caption{IAA for all annotated dimensions.}\label{tab:iaa}
\end{table*}

\section{Label Distribution}\label{app:label_dis}

\begin{figure}[h]
    \centering
    \includegraphics[width=\linewidth]{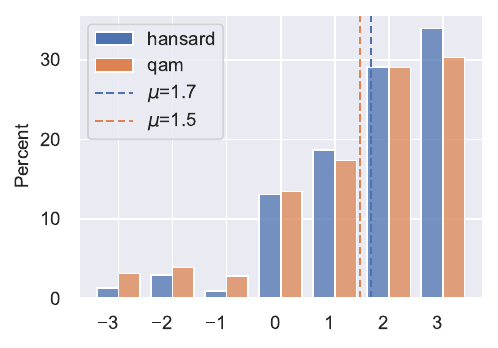}
    \caption{Distribution of \logic{} ratings.}
    \label{fig:dis_context}
\end{figure}

\begin{figure}
    \centering
    \includegraphics[width=\linewidth]{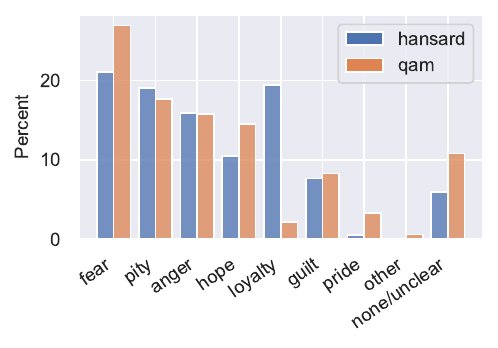}
    \includegraphics[width=.93\linewidth]{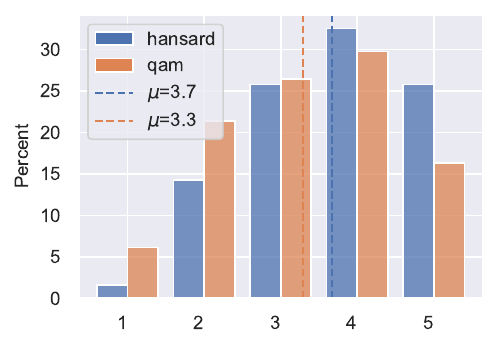}
    \caption{Distribution of \eat{} (top) and \eas{} (bottom).}
    \label{fig:dis_ea}
\end{figure}

\begin{figure}[!ht]
    \centering
    \includegraphics[width=\linewidth]{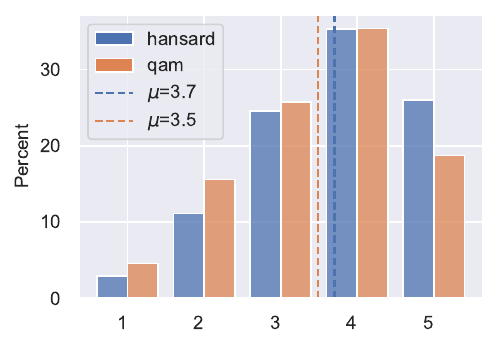}
    \includegraphics[width=\linewidth]{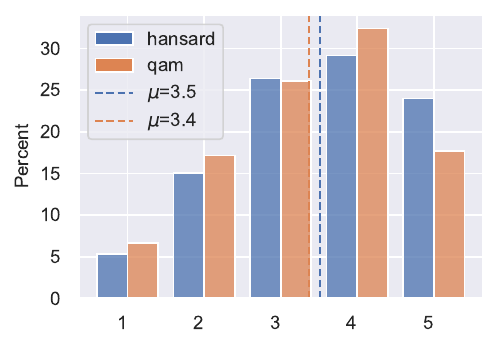}
    \includegraphics[width=\linewidth]{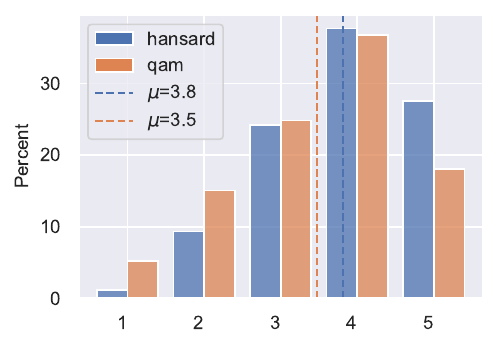}
    \caption{From top to bottom: distributions of \fit{}, \evi{}, and \norm{} ratings.}
    \label{fig:dis_ea_emo}
\end{figure}

\begin{figure}[!ht]
    \centering
    \includegraphics[width=.93\linewidth]{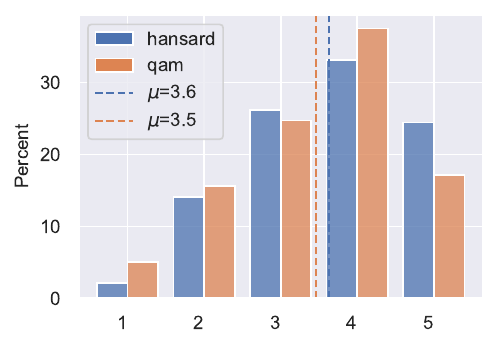}
    \includegraphics[width=\linewidth]{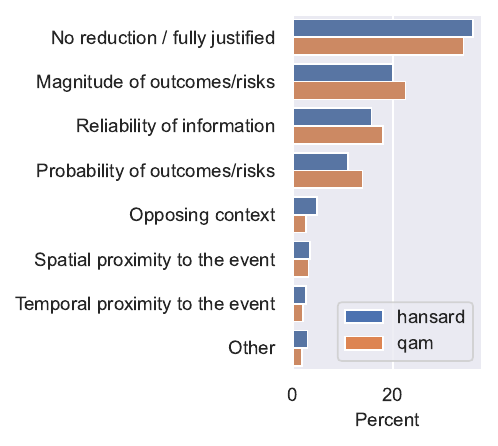}
    \caption{Top: Distribution of overall EA justification ratings. Bottom: Distribution of reasons selected for reduction in EA evidential justification scores.}
    \label{fig:dis_ea_jus}
\end{figure}

Figure~\ref{fig:dis_context} shows the distribution of \logic{} ratings. Figure~\ref{fig:dis_ea} shows the distributions of the \eat{} and \eas{} labels. Figure~\ref{fig:dis_ea_emo} shows the distribution of \fit{}, \evi{}, and \norm{} ratings. Figure~\ref{fig:dis_ea_jus} shows the distribution of the overall justification ratings and that of the reasons for reductions in \evi{}.

\section{Results}
\begin{figure}[!ht]
    \centering
    \includegraphics[width=.495\linewidth]{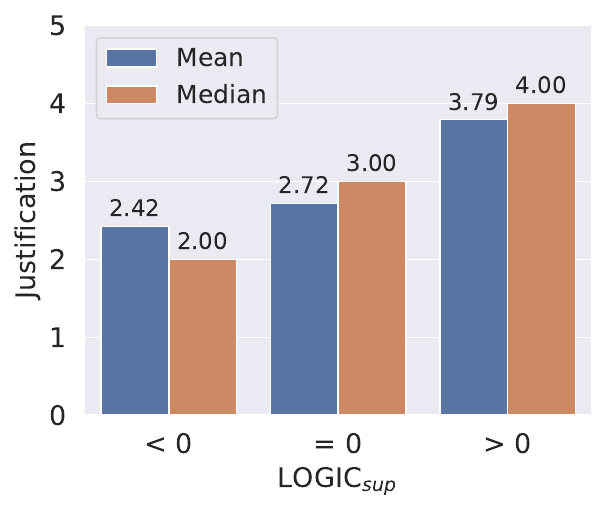}
    \includegraphics[width=.495\linewidth]{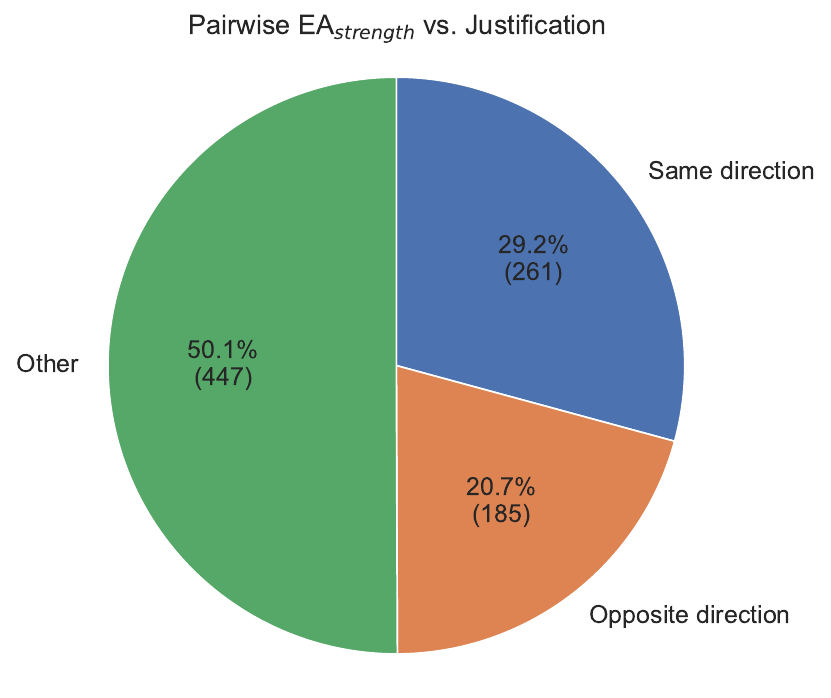}
    \caption{Left: Average EA justification ratings for opposing, neutral, and supporting contexts. Right: Percentages and counts of pairwise comparisons between two items with the same context, where \eas{} and justification ratings change in the same direction, in opposite directions, or neither.}
    \label{fig:bar_pie}
\end{figure}

\begin{figure}[!ht]
    \centering
    \includegraphics[width=\linewidth]{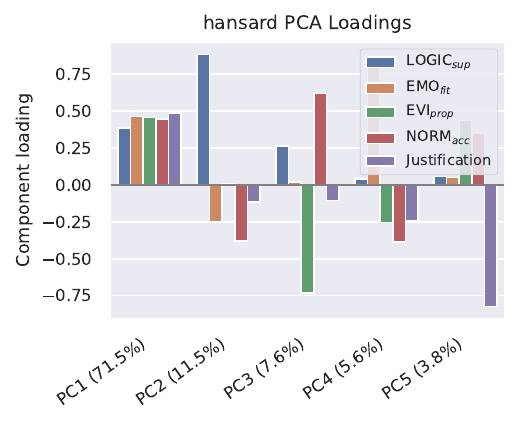}
    \includegraphics[width=\linewidth]{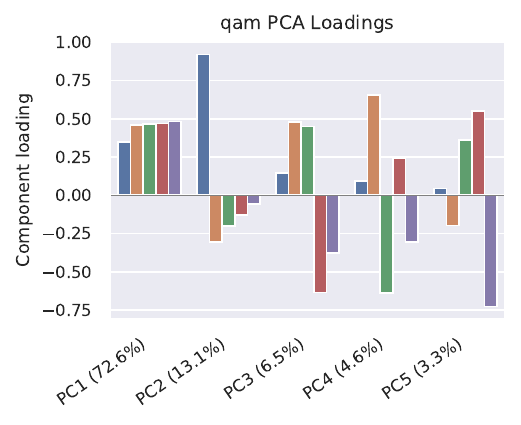}
    \caption{PCA loadings for each dataset.}
    \label{fig:pca}
\end{figure}

Figure~\ref{fig:bar_pie} shows verage EA justification ratings for
opposing, neutral, and supporting contexts, and Percentages and counts of pairwise comparisons between two items with the same context, where \eas{} and
justification ratings change in the same direction, in opposite directions, or neither. Figure~\ref{fig:pca} shows the PCA loadings per dataset.

\begin{table*}[t]
\centering
\small
\setlength{\tabcolsep}{3pt}
\renewcommand{\arraystretch}{1.02}
\resizebox{\textwidth}{!}{%
\begin{tabular}{@{}>{\raggedright\arraybackslash}p{6.2cm}>{\raggedright\arraybackslash}p{5.0cm}>{\raggedright\arraybackslash}p{2.7cm}>{\raggedright\arraybackslash}p{3.2cm}@{}}
\toprule
\textbf{Topic \& Argument} & \textbf{Context} & \textbf{Dimension} & \textbf{Top-3 majority label} \\
\midrule
\multirow[t]{8}{6.2cm}{\texttt{qam\_0391}\par\smallskip \textit{Topic:} Should animal testing be banned\par\smallskip Thus, because animals are subjected to agonizing pain, suffering and death when they are used in laboratory and cosmetics testing, animal research must be stopped to prevent more waste of animal life.} &
\multirow[t]{8}{5.0cm}{According to Orlans, the animals suffer from mild discomfort and temporary loss of appetite.} &
\logic{} & -3 \\
\cmidrule(lr){3-4}
& & \eat{} & pity / compassion / sympathy \\
\cmidrule(lr){3-4}
& & \eas{} & 5 \\
\cmidrule(lr){3-4}
& & \fit{} & 1 \\
\cmidrule(lr){3-4}
& & \evi{} & 1 \\
\cmidrule(lr){3-4}
& & Reasons & Magnitude of outcomes/risks \\
\cmidrule(lr){3-4}
& & \norm{} & 1 \\
\cmidrule(lr){3-4}
& & Overall & \textbf{1} \\
\midrule
\multirow[t]{8}{6.2cm}{\texttt{qam\_0375}\par\smallskip \textit{Topic:} Will artificial intelligence replace humans\par\smallskip Highly capable and autonomous planning systems require additional checks because of their potential to generate plans that treat humans adversarially, as competitors for limited resources .} &
\multirow[t]{8}{5.0cm}{There are some goals that very few artificial intelligences might rationally pursue, like acquiring additional resources or self-preservation . This could prove problematic because it might put an artificial intelligence in direct competition with humans.} &
\logic{} & 1 \\
\cmidrule(lr){3-4}
& & \eat{} & fear / threat \\
\cmidrule(lr){3-4}
& & \eas{} & 2 \\
\cmidrule(lr){3-4}
& & \fit{} & 1 \\
\cmidrule(lr){3-4}
& & \evi{} & 1 \\
\cmidrule(lr){3-4}
& & Reasons & Magnitude of outcomes/risks; Probability of outcomes/risks; Reliability of information \\
\cmidrule(lr){3-4}
& & \norm{} & 3 \\
\cmidrule(lr){3-4}
& & Overall & \textbf{1} \\
\midrule
\multirow[t]{8}{6.2cm}{\texttt{qam\_0044}\par\smallskip \textit{Topic:} Should animal testing be banned\par\smallskip Thus, because animals are subjected to agonizing pain, suffering and death when they are used in laboratory and cosmetics testing, animal research must be stopped to prevent more waste of animal life.} &
\multirow[t]{8}{5.0cm}{The LD50 test is used to test the dosage of a substance that is necessary to cause death in fifty percent of the animal subjects within a certain amount of time. To perform this test, the researchers hook the animals up to tubes that pump huge amounts of the test product into their stomachs until they die. This test is extremely painful to the animals because death can take days or even weeks.} &
\logic{} & 3 \\
\cmidrule(lr){3-4}
& & \eat{} & pity / compassion / sympathy \\
\cmidrule(lr){3-4}
& & \eas{} & 5 \\
\cmidrule(lr){3-4}
& & \fit{} & 5 \\
\cmidrule(lr){3-4}
& & \evi{} & 5 \\
\cmidrule(lr){3-4}
& & Reasons & No reduction / fully justified \\
\cmidrule(lr){3-4}
& & \norm{} & 5 \\
\cmidrule(lr){3-4}
& & Overall & \textbf{5} \\
\midrule
\multirow[t]{8}{6.2cm}{\texttt{qam\_0040}\par\smallskip \textit{Topic:} Will artificial intelligence replace humans\par\smallskip Highly capable and autonomous planning systems require additional checks because of their potential to generate plans that treat humans adversarially, as competitors for limited resources .} &
\multirow[t]{8}{5.0cm}{There are some goals that almost any artificial intelligence might rationally pursue, like acquiring additional resources or self-preservation . This could prove problematic because it might put an artificial intelligence in direct competition with humans.} &
\logic{} & 3 \\
\cmidrule(lr){3-4}
& & \eat{} & fear / threat \\
\cmidrule(lr){3-4}
& & \eas{} & 2 \\
\cmidrule(lr){3-4}
& & \fit{} & 5 \\
\cmidrule(lr){3-4}
& & \evi{} & 5 \\
\cmidrule(lr){3-4}
& & Reasons & No reduction / fully justified \\
\cmidrule(lr){3-4}
& & \norm{} & 5 \\
\cmidrule(lr){3-4}
& & Overall & \textbf{5} \\
\bottomrule
\end{tabular}%
}
\caption{Selected examples in QAM with low ($\leq 2$) and high ($\geq 4$) overall emotional-appeal justification. }
\label{tab:ex_qam}
\end{table*}

\begin{table*}[t]
\centering
\small
\setlength{\tabcolsep}{3pt}
\renewcommand{\arraystretch}{1.02}
\resizebox{\textwidth}{!}{%
\begin{tabular}{@{}>{\raggedright\arraybackslash}p{6.2cm}>{\raggedright\arraybackslash}p{5.0cm}>{\raggedright\arraybackslash}p{2.7cm}>{\raggedright\arraybackslash}p{3.2cm}@{}}
\toprule
\textbf{Topic \& Argument} & \textbf{Context} & \textbf{Dimension} & \textbf{Top-3 majority label} \\
\midrule
\multirow[t]{8}{6.2cm}{\texttt{hansard\_0333}\par\smallskip \textit{Topic:} Illegal migration should be stopped.\par\smallskip If we fail to deter illegal ways of reaching our shores, how long before the security of every town and city in this country is compromised?} &
\multirow[t]{8}{5.0cm}{The Royal National Lifeboat Institution recalled rescuing children with chemical burns from leaked petrol, and others so frozen that they could not walk because they had spent 30 hours at sea.} &
\logic{} & 0 \\
\cmidrule(lr){3-4}
& & \eat{} & fear / threat \\
\cmidrule(lr){3-4}
& & \eas{} & 3 \\
\cmidrule(lr){3-4}
& & \fit{} & 1 \\
\cmidrule(lr){3-4}
& & \evi{} & 1 \\
\cmidrule(lr){3-4}
& & Reasons & Magnitude of outcomes/risks; Probability of outcomes/risks; Reliability of information \\
\cmidrule(lr){3-4}
& & \norm{} & 1 \\
\cmidrule(lr){3-4}
& & Overall & \textbf{1} \\
\midrule
\multirow[t]{8}{6.2cm}{\texttt{hansard\_0197}\par\smallskip \textit{Topic:} Aid must be allowed into Gaza.\par\smallskip Mothers in Gaza are watching their children waste away, unable to find food, unable to find medicine, unable to find even clean water. If we have any shred of compassion left in this House, we must ensure that aid reaches Gaza now.} &
\multirow[t]{8}{5.0cm}{The peace plan said that a minimum of 600 trucks a day would be allowed into Gaza to support more than 2 million people to meet their basic needs.} &
\logic{} & 1 \\
\cmidrule(lr){3-4}
& & \eat{} & pity / compassion / sympathy \\
\cmidrule(lr){3-4}
& & \eas{} & 4 \\
\cmidrule(lr){3-4}
& & \fit{} & 2 \\
\cmidrule(lr){3-4}
& & \evi{} & 2 \\
\cmidrule(lr){3-4}
& & Reasons & Probability of outcomes/risks; Reliability of information \\
\cmidrule(lr){3-4}
& & \norm{} & 3 \\
\cmidrule(lr){3-4}
& & Overall & \textbf{2} \\
\midrule
\multirow[t]{8}{6.2cm}{\texttt{hansard\_0271}\par\smallskip \textit{Topic:} Illegal migration should be stopped.\par\smallskip We need to stop these dangerous crossings and restore order to a broken asylum system.} &
\multirow[t]{8}{5.0cm}{The Royal National Lifeboat Institution recalled rescuing children with chemical burns from leaked petrol, and others so frozen that they could not walk because they had spent 30 hours at sea.} &
\logic{} & 3 \\
\cmidrule(lr){3-4}
& & \eat{} & fear / threat \\
\cmidrule(lr){3-4}
& & \eas{} & 4 \\
\cmidrule(lr){3-4}
& & \fit{} & 4 \\
\cmidrule(lr){3-4}
& & \evi{} & 4 \\
\cmidrule(lr){3-4}
& & Reasons & Other \\
\cmidrule(lr){3-4}
& & \norm{} & 4 \\
\cmidrule(lr){3-4}
& & Overall & \textbf{4} \\
\midrule
\multirow[t]{8}{6.2cm}{\texttt{hansard\_0199}\par\smallskip \textit{Topic:} Aid must be allowed into Gaza.\par\smallskip We must do whatever it takes to get aid into Gaza to stop people dying of starvation} &
\multirow[t]{8}{5.0cm}{The peace plan said that a minimum of 600 trucks a day would be allowed into Gaza to support more than 2 million people to meet their basic needs.} &
\logic{} & 3 \\
\cmidrule(lr){3-4}
& & \eat{} & pity / compassion / sympathy \\
\cmidrule(lr){3-4}
& & \eas{} & 5 \\
\cmidrule(lr){3-4}
& & \fit{} & 5 \\
\cmidrule(lr){3-4}
& & \evi{} & 5 \\
\cmidrule(lr){3-4}
& & Reasons & No reduction / fully justified \\
\cmidrule(lr){3-4}
& & \norm{} & 5 \\
\cmidrule(lr){3-4}
& & Overall & \textbf{5} \\
\bottomrule
\end{tabular}%
}
\caption{Selected examples in Hansard with low ($\leq 2$) and high ($\geq 4$) overall emotional-appeal justification.}\label{tab:ex_hansard}
\end{table*}

\end{document}